\documentclass{article}
\usepackage{iclr2027_conference,times}

\usepackage{amsmath,amsfonts,bm}

\def\eqref#1{equation~\ref{#1}}

\def\1{\bm{1}}

\DeclareMathAlphabet{\mathsfit}{\encodingdefault}{\sfdefault}{m}{sl}
\SetMathAlphabet{\mathsfit}{bold}{\encodingdefault}{\sfdefault}{bx}{n}

\usepackage{amsmath,amssymb,booktabs,xcolor,xspace,hyperref,url,graphicx}
\usepackage{algorithm,algpseudocode,listings}
\usepackage[breakable,skins]{tcolorbox}
\newcounter{paperprompt}
\lstdefinestyle{paperprompt}{basicstyle=\normalsize\ttfamily,
  columns=fullflexible,keepspaces=true,breaklines=true,breakatwhitespace=true,
  showstringspaces=false,showtabs=false,tabsize=2,aboveskip=2pt,belowskip=2pt}
\tcbset{paperpromptbox/.style={enhanced,breakable=false,colback=black!2,colframe=black!35,
  colbacktitle=black!7,coltitle=black,fonttitle=\normalsize\bfseries,
  boxrule=0.4pt,arc=0pt,left=6pt,right=6pt,top=5pt,bottom=5pt,
  before skip=8pt,after skip=8pt}}
\newenvironment{promptbox}[2]{%
  \refstepcounter{paperprompt}\label{#2}%
  \tcolorbox[paperpromptbox,title={Prompt \thepaperprompt: #1}]%
  \begingroup\normalfont\normalsize}{%
  \endgroup\endtcolorbox}

\hypersetup{hidelinks}
\title{Towards Communication-Efficient\\Social Intelligence in Language Agents}
\ifdefined\TACTreviewcopy
  \author{Anonymous Authors}
\else
\newcommand{\TACTauthorblock}{%
\begin{minipage}[t]{0.97\textwidth}
\centering\normalfont\normalsize
Linxiao Gong\textsuperscript{1,*}, Yijie Xu\textsuperscript{1,*},
Tianfu Wang\textsuperscript{1,*,$\ddagger$}, Yin Wu\textsuperscript{1}, Yili Wang\textsuperscript{1},\\
Xingbo Yao\textsuperscript{1}, Huizai Yao\textsuperscript{1},
Xilin Xia\textsuperscript{2}, Haowen Yang\textsuperscript{1}, Hui Xiong\textsuperscript{3,\dag}
\par\vspace{0.65em}
{\small
\textsuperscript{1}The Hong Kong University of Science and Technology (Guangzhou)\\
\textsuperscript{2}University of Science and Technology of China\\
\textsuperscript{3}The Hong Kong University of Science and Technology\par}
\vspace{0.45em}
{\footnotesize
\textsuperscript{*}Equal contribution\quad
\textsuperscript{$\ddagger$}Project leader\quad
\textsuperscript{\dag}Corresponding author \\
\href{mailto:tianfuwang.cs@gmail.com}{\texttt{tianfuwang.cs@gmail.com}} \quad
\href{mailto:xionghui@ust.hk}{\texttt{xionghui@ust.hk}}\par}
\end{minipage}%
}
\author{\TACTauthorblock}
\hypersetup{
  pdftitle={Towards Communication-Efficient Social Intelligence in Language Agents},
  pdfauthor={Linxiao Gong, Yijie Xu, Tianfu Wang, Yin Wu, Yili Wang, Xingbo Yao, Huizai Yao, Xilin Xia, Haowen Yang, Hui Xiong}
}

  \iclrfinalcopy
\fi
\begin{document}
\maketitle
\ifdefined\TACTreviewcopy\else
  \lhead{Preprint}
\fi
\raggedbottom
\begin{abstract}
Socially intelligent language agents must negotiate, coordinate, and resolve
conflicting preferences while respecting the time and attention of both
participants. Balancing these demands is challenging because agents must convey
enough to address a partner's constraints and advance their goals without adding
words that do not help the interaction. In this paper, we propose
Teacher-Assisted Communication Training (TACT) to improve social goal attainment
while reducing communication cost, making interactions with agents more
productive and less demanding. We first characterize communication efficiency
in terms of action strategy and expression, whose effects extend beyond the
current utterance to the partner's response and subsequent exchanges. We design
TACT to revise student-generated actions, test the revisions through partner
responses, and distill useful feedback into the student. An expression
specialist removes unnecessary detail while preserving the intended action,
while a strategy specialist proposes alternatives that may better address the
partner's constraints. To determine which revision helps, TACT samples a partner
response for each candidate and selects a teacher reference by balancing local
goal support against action-token cost. That reference guides on-policy
distillation on the student's own generation prefixes, allowing the student to
act independently at deployment. We evaluate TACT on SOTOPIA and AgentSense. On
SOTOPIA, it achieves the highest Goal among the evaluated methods on All and
Hard while using substantially fewer target tokens than SFT+SDPO. On
AgentSense, it improves goal success over the initial student while reducing
target tokens and interaction messages.
\end{abstract}

\section{Introduction}
Language agents use communication to negotiate, coordinate plans, and navigate
conflicting preferences~\citep{lewis2017negotiation,park2023generative,zhou2024sotopia}.
Social intelligence in these settings requires understanding others'
constraints and choosing appropriate actions as the situation develops. An
agent's actions shape what the partner understands and how they respond, which
in turn affects whether the conversation moves forward or requires clarification.
These exchanges take time and attention from both participants as they establish
enough common ground to proceed~\citep{clark1991grounding}; two interactions can
therefore reach the same goal while requiring very different amounts of
communication. We study communication-efficient social interaction by
considering goal attainment together with the communication used to achieve it.

\begin{figure}[t]
  \centering
  \vspace{4pt}
  \includegraphics[width=\linewidth]{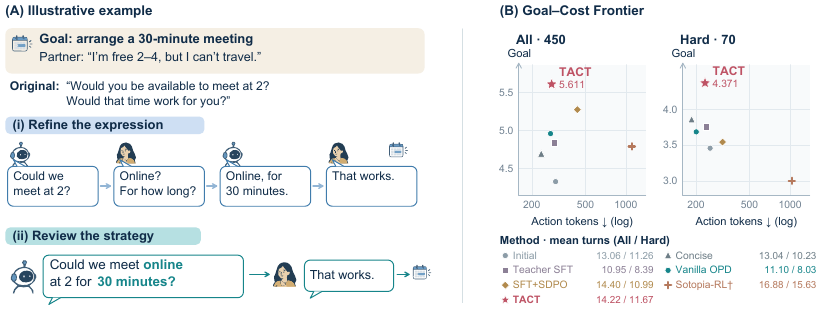}
  \vspace{-7pt}
  \caption{\textbf{Communication efficiency depends on the whole interaction.}
  (A) A shorter request can require additional clarification, whereas a more
  informative strategy revision reaches the same goal in fewer exchanges.
  Dialogues are illustrative, not experimental outputs.
  (B) TACT lies on the empirical Goal--Cost Pareto frontier among the methods
  shown on both SOTOPIA-All and Hard.}
  \label{fig:motivation}
  \vspace{4pt}
\end{figure}

The communication needed to reach a goal is not captured by the length of any
one response (Figure~\ref{fig:motivation}). As participants establish common
ground, descriptions that require detail early in an exchange can become
shorter later~\citep{clark1986referring}. Yet shorter wording can also leave out
information the partner needs, and it cannot repair an action that ignores a
stated constraint. An agent must therefore choose both an action that can move
the interaction forward and an expression suited to what the partner already
knows. The partner's response provides an early indication of whether these
choices help; their overall value depends on goal attainment and communication
over the complete interaction.

Turning these interaction-level judgments into training signals raises three
linked challenges. First, feedback must lead to a concrete revision of the
student's current action. Rated social dialogues support social-agent
learning~\citep{wang2024sotopiapi}, and utterance-level rewards improve credit
assignment~\citep{yu2025sotopiarl}, but neither alone specifies what the student
should change. Second, a proposed revision must be assessed through interaction
before it becomes supervision. Simulated continuations can probe its effect on
subsequent responses~\citep{wu2025collabllm}, while choosing among revisions
requires weighing goal progress against communication cost. Finally, the
selected guidance must reach the student's own generation. Training only on
revised responses can miss the prefixes the student encounters when acting
independently~\citep{ross2011dagger,agarwal2024gkd}.

We introduce \textbf{TACT} (Teacher-Assisted Communication Training) to improve
social goal attainment without unnecessary communication. TACT begins with an
action the student actually generated, so that supervision addresses a decision
the student faced. An expression specialist proposes a more concise version
while preserving intent and commitments; a strategy specialist can instead
revise the action to address the partner's constraints. Neither proposal is
accepted on appearance alone: shorter wording may omit needed information,
while a different action may still fail to move the exchange forward. TACT
therefore samples a partner response to the original action and each proposal,
then selects a reference using a local goal-support proxy and action-token cost.
For this guidance to help when the student acts alone, the selected reference
is given only to the teacher during on-policy distillation
(OPD)~\citep{agarwal2024gkd}. The teacher provides feedback on the student's
original token prefixes, while the student retains its own input context and
needs no specialist at deployment.

We evaluate TACT on SOTOPIA and AgentSense. On SOTOPIA, it achieves the highest
Goal among the evaluated methods on both All and Hard, while using
substantially fewer target tokens than SFT+SDPO. On AgentSense, it improves goal
success over the initial student while reducing target tokens and interaction
messages. Component comparisons show that expression and strategy supervision
affect token use and interaction turns differently.

We summarize our main contributions as follows:
\begin{itemize}
\item We characterize communication efficiency beyond response length by
identifying two sources of avoidable cost: excess wording within an action, and
further exchanges that can follow when the action leaves a partner's constraints
unresolved.
\item We propose TACT, which revises student actions through expression and
strategy specialists, tests the revisions through partner responses, and uses
goal support and token cost to select a reference for on-policy distillation.
\item We evaluate TACT on SOTOPIA and AgentSense, finding stronger goal
attainment with fewer target tokens than the initial student in both. Ablations
show distinct effects of expression and strategy supervision on token use and
interaction turns.
\end{itemize}

\section{Related Work}
\noindent\textbf{Social Intelligence.}
Research on social intelligence in LLMs connects evaluation and applications
with learning from interaction. Interactive benchmarks assess goal attainment,
relationship change, and implicit information reasoning~\citep{zhou2024sotopia,mou2025agentsense},
while agentic social-skill tutoring supports human learning through situated
practice and reflective feedback~\citep{wang2026socialcoach}. To improve agents'
own behavior, another line learns from interaction data through behavior
cloning, self-reinforcement, strategy injection, preference optimization, and
reinforcement learning~\citep{wang2024sotopiapi,zhang2025sotopiaomega,kong2025sdpo,yu2025sotopiarl}.
Communication efficiency also receives attention: ASL combines adaptive
reasoning-mode selection with an answer-length reward to improve goal
attainment and token efficiency~\citep{wang2026adaptivesocial}. TACT studies
expression and strategy revisions as training references for improving
communication over complete interactions.

\noindent\textbf{Interaction-Based Supervision.}
Learning from interaction requires connecting feedback to individual actions.
Existing work refines dialogue-level supervision through reward decomposition
and contribution estimation, and evaluates intermediate actions through process
feedback, expected utility over future interactions, and information
gain~\citep{yu2025sotopiarl,feng2026savoir,wang2026socialrl,wang2026igpo}.
Simulated continuations also support cost-aware supervision: CollabLLM combines
task success, communication cost, and engagement in multi-turn rewards~\citep{wu2025collabllm}.
TACT samples one partner response for each original or candidate action, then
compares local goal support and token cost under the same context to select a
distillation reference.

\noindent\textbf{On-Policy Distillation.}
On-policy distillation converts teacher judgments into supervision at
student-generated prefixes~\citep{agarwal2024gkd}. Recent work enriches teacher
contexts with demonstrations, correct solutions, environment feedback, and
subsequent user messages to produce token-level
supervision~\citep{shenfeld2026sdft,zhao2026opsd,hubotter2026selfdistillation,buening2026userinteractions}.
Related work extracts hindsight skills from completed trajectories to support
policy updates~\citep{wu2026seed}. TACT supplies interaction-selected revisions
only to the teacher, which provides feedback on the student's original
prefixes; the student acts independently at deployment.

\section{Method}
\label{sec:method}
\begin{figure}[!t]
  \centering
  \includegraphics[width=\linewidth]{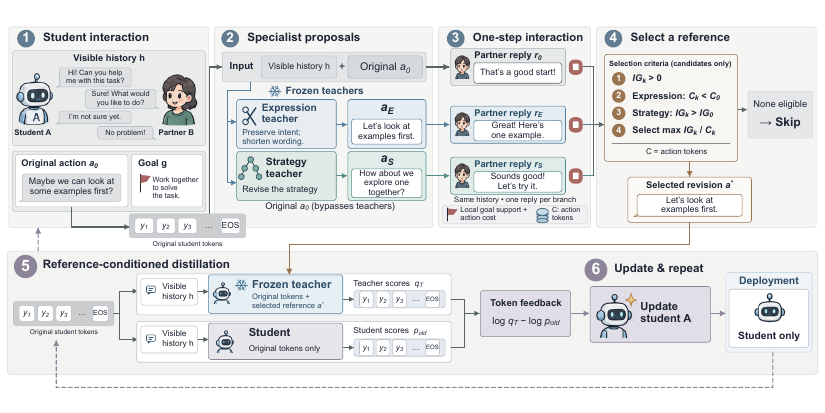}
  \caption{\textbf{Overview of TACT.} Expression and strategy specialists
  propose revisions to student actions. TACT selects a reference using local
  interaction feedback and token cost, then conditions the teacher on this
  reference to guide on-policy distillation.}
  \label{fig:pipeline}
\end{figure}

As shown in Figure~\ref{fig:pipeline}, TACT first uses expression and strategy
specialists to propose complementary revisions to student actions under the
same visible context.
It then compares the original action and candidate revisions through one
partner reply per branch, selecting an eligible reference based on goal-support
gain per action token.
Finally, the selected reference provides additional context to the teacher,
whose token-level feedback on the student's original action guides on-policy
distillation and subsequent rounds of interaction.

\subsection{Problem Formulation}
\label{sec:formulation}
We study communication efficiency in multi-turn social interactions by
considering goal attainment together with the communication required to
achieve it. Expression and action choice jointly determine what an agent
communicates as an interaction unfolds. At turn $t$, an agent's visible context $h_t$
contains the scenario and role information, its own goal, and the dialogue
history observed so far. The student policy samples an action
$a_t\sim\pi_\theta(\cdot\mid h_t)$, and successive exchanges with its partner
form a complete interaction trajectory $\tau$.

We measure interaction outcomes by the final goal attainment score $G(\tau)$
and communication costs by action tokens and interaction turns. Let $C(a_t)$
denote the token length of action $a_t$. The target agent's total action-token
count is
\begin{equation}
 N_{\mathrm{action}}(\tau)
 =\sum_{t\in\mathcal I(\tau)} C(a_t),
\label{eq:action-cost}
\end{equation}
where $\mathcal I(\tau)$ indexes the target agent's actions in the trajectory.
We use $N_{\mathrm{turn}}(\tau)$ to denote the total number of environment
turns taken by both participants. These costs capture the target agent's
language output and the exchanges required to complete the interaction,
respectively.

\subsection{Dual-Specialist Revision}
\label{sec:diagnosis}
\paragraph{On-Policy Rollouts.}
Collect two-role interactions using the same student snapshot for both roles.
Only the designated learning slot A contributes training loss; B acts as its
partner. Retain each action's visible history, exact generated tokens, ending
markers, behavior probabilities, sampling settings, and termination status.
Both roles refresh with the student between batches. The completed experiments use explicit
non-thinking model templates; stripping generated reasoning afterward would not
be an equivalent procedure. Fixed template boundaries, if required by the model,
remain input context rather than supervised output.

\paragraph{Specialist Editing Permissions.}
For every A action, expression and strategy specialists each generate one
alternative. There is no probability probe, preselected subset of turns, or
quality-driven resampling. Both receive the target-time role-visible history
and original action, but not its recorded future. They can share frozen model
parameters while retaining different instructions.

The \emph{expression specialist} improves clarity and wording while preserving
the original action's intent, facts, commitments, and action type. It must not
invent facts, add or withdraw commitments, or change the negotiation strategy.
The \emph{strategy specialist} can change intent and action type: for example,
accepting a stated constraint, offering another arrangement, seeking confirmation,
or ending the interaction. It must still respect visible facts and boundaries.
These are complementary editing permissions, not separate cost definitions.
Expression candidates must be shorter to become eligible; strategy candidates
can be longer when they improve the local goal-support measure. Prompt
instructions specify semantic preservation but do not implement a verified
semantic-consistency filter. Candidate outputs therefore remain available for semantic-compliance inspection.

The original trajectory stays intact. Later target actions still come from the
student's collected dialogue, not from stitched candidate branches. Candidate
construction, response comparison, and learning therefore preserve distinct
records and information conditions.

\subsection{Efficiency-Aware Selection}
\label{sec:validation}
\paragraph{Single-Response Branches.}
Treat the original action as branch $0$ and specialist alternatives as branches
$k\in\{1,2\}$. Starting from the same public history, each receives one new
legal partner response $r_t^k$, then stops. Use the same partner snapshot and generation settings, with one recorded
partner seed shared across distinct actions at the same node. Different prompts
still produce different conditional responses.
The partner sees its own role information and the branch's outward action,
not teacher identity, candidate-generation instructions, or the original future. The original
branch also receives a newly sampled reply rather than reusing its recorded
reply. Recorded seeds support reproducibility but do not remove response uncertainty.

If the original branch cannot obtain a legal, scorable response, skip this
target action. A failed candidate branch only excludes that candidate. Leaving
or other terminal actions do not receive fabricated partner responses. The
method does not continue branches to the end of the dialogue.

\paragraph{Local Goal Support.}
Let $g=(g_1,\ldots,g_m)$ be the original role goal text, without an added achievement
frame, paraphrase, or decomposition. For a scoring context $H$, define
\begin{equation}
 s_b(H,g)=\frac{1}{m}\sum_{j=1}^{m}
 \log\pi_{\theta_b}(g_j\mid \mathcal T(H),g_{<j}),
\label{eq:support}
\end{equation}
where $\mathcal T$ is a fixed non-thinking scoring template with an explicit
answer boundary. Only target-text tokens are scored. The scorer
is the batch student snapshot, frozen within the batch and refreshed between
batches; no independent candidate-quality model is introduced.

Let $H_t^-$ be the learning role's context before its action and $H_t^{k,+}$
include its branch outward action and received public response.
Neither contains partner-private information. Define
\begin{align}
 \mathrm{IG}_t^k &= s_b(H_t^{k,+},g)-s_b(H_t^-,g),\\
 G_t^k &= \mathrm{IG}_t^k-\mathrm{IG}_t^0.
\label{eq:gain}
\end{align}
The shared pre-action term cancels in the candidate--original difference.
This goal-support change adapts the use of answer-probability changes in
IGPO~\citep{wang2026igpo}; here the target is a social goal statement, not a
known correct search answer. The score captures the information gain contributed by a reply under the current dialogue context. It is designed to identify responses that provide meaningful evidence toward goal-relevant social progress.

\paragraph{Efficiency-Aware Selection.}
Write $I_k=\mathrm{IG}_t^k$, $I_0=\mathrm{IG}_t^0$, and $C_k=C(a_t^k)$.
Consider only legal, scorable alternatives that differ from the original action.
Both specialists require $I_k>0$. An expression candidate additionally requires
$C_k<C_0$; a strategy candidate additionally requires $I_k>I_0$.
An expression candidate can therefore be eligible even when its goal-support
change is below the original action's. These are proxy-based rules, not a
non-inferiority test for final social quality.

Let $\mathcal K_t$ contain the eligible candidates and $\mathcal F_t$ its
nondominated subset under larger $I_k$ and smaller $C_k$. Select
\begin{equation}
 k_t^*\in\arg\max_{k\in\mathcal F_t}\frac{I_k}{C_k}.
\label{eq:selection}
\end{equation}
No eligible candidate means no update from this action; exact ties use a
recorded seeded draw. All costs include the serialized action and required
ending tokens. The relative gain $G_t^k$ remains a diagnostic and supplies the
strategy eligibility check; it is not the numerator of the selection ratio.
On positive scores and costs, ratio maximization is Pareto-consistent, so
frontier filtering is not a separate algorithmic contribution.

Distinct legal actions each receive one partner response; identical actions
share their branch result. Termination without a partner response remains
unscored. This local comparison is a contribution proxy, not a causal attribution
of the full dialogue outcome.

Save the selected specialist, candidate/control records, local scores, costs,
and exclusion reasons. Selection does not label a candidate socially sufficient.
For example, a concise commitment may raise goal support after a cooperative
reply while leaving the agreement ambiguous. The proxy can miss this failure,
especially when its consequences occur later.

\subsection{On-Policy Distillation}
\label{sec:distillation}
Following \citet{agarwal2024gkd}, we train the student using teacher feedback
at student-generated action prefixes. In TACT, the selected revision serves
as additional teacher context: the teacher evaluates the student's original
action tokens with this reference, while the student retains its original
visible context. The reference guides token-level feedback without replacing
the student's action or exposing the candidate branch's subsequent reply
to the scoring teacher.

We use sampled-token teacher--student log-probability differences as
feedback~\citep{thinkingmachines2025opd}. After each batch update, the refreshed
student collects new interactions; both specialists and the scoring teacher
remain frozen. At deployment, the student acts independently with its original
input. The feedback formula and optimization details are provided in
Appendix~\ref{app:actor-update}.

\section{Experiments}
\label{sec:experiments}
We evaluate whether TACT improves social goal attainment while reducing
communication cost on SOTOPIA and AgentSense.
The code will be released upon acceptance of the paper.

\subsection{Experimental Setup}

\noindent \textbf{Datasets.}
We evaluate on SOTOPIA~\citep{zhou2024sotopia}, where agents pursue private social goals through multi-turn interaction. Training collects 16 online dialogues per iteration from a seeded randomized schedule, using scenarios separate from evaluation. The evaluation panel contains 90 scenarios with five character pairings each (450 settings), including the official 14-scenario SOTOPIA-Hard subset. A fixed 90-case panel supports development analyses. We additionally evaluate transfer to 500 AgentSense instances from 100 fixed two-person templates~\citep{mou2025agentsense}, using a separate goal/relationship protocol (Appendix~\ref{app:agentsense}).

\noindent \textbf{Baselines.}
We compare with the \textbf{initial student}, a \textbf{concise-prompt baseline}, \textbf{Teacher SFT}, and \textbf{vanilla OPD}. The latter two use teacher demonstrations and on-policy token feedback, respectively, without specialist selection or selected-reference conditioning. \textbf{Prompted OPD} adds a teacher-only instruction to jointly improve goal progress, strategy, and conciseness to the vanilla OPD recipe, without candidate selection or reference conditioning. \textbf{SFT+SDPO}~\citep{kong2025sdpo} applies segment-preference optimization after public-data SFT. \textbf{Sotopia-RL}~\citep{yu2025sotopiarl} uses the published Qwen2.5-7B checkpoint as an external-backbone reference. Historical \textbf{same-context OPD} retains selection but omits the teacher's selected reference. Training details appear in Appendix~\ref{app:baseline-training}.

\noindent \textbf{Models.}
We use \textbf{Qwen3.5-4B} as the student and \textbf{Qwen3.5-27B} as the frozen backbone for both specialists, distinguished by their prompts. Only the target agent is trained, using LoRA. The training partner shares the batch's student snapshot; the evaluation partner is the initial Qwen3.5-4B. The batch-frozen student snapshot also computes IG; the 27B teacher supplies reference-conditioned token probabilities for distillation. Native thinking is disabled. \textbf{DeepSeek-v4-pro} judges completed dialogues using the SOTOPIA rubric; these scores are not training rewards. Optimization settings and prompts appear in Appendices~\ref{app:implementation} and~\ref{app:prompts}.

\noindent \textbf{Evaluation.}
We report the seven SOTOPIA dimensions and their native-scale mean (Avg), cumulative target-agent action \textbf{Tokens}, and both participants' interaction \textbf{Turns}. All and Hard contain 450 and 70 settings, with matched scenarios, character pairings, role assignments, and evaluation partner. All populated main and specialist rows cover these complete panels after recorded recovery. Decoding settings appear in Appendix~\ref{app:implementation}; recovery procedures, metric definitions, and uncertainty calculations appear in Appendix~\ref{app:evaluation-details}.

\subsection{Results}
\label{sec:results}
\noindent\textbf{Social Performance.}
TACT offers a favorable balance between social goal attainment and token cost, achieving the highest Goal score among the evaluated methods on both SOTOPIA-All and SOTOPIA-Hard (Table~\ref{tab:main-results}). Its advantage extends from the initial student and concise prompting to demonstration-based and on-policy training baselines. The same pattern on Hard shows that the improvement also appears in the benchmark's more challenging interactions. Full social scores, paired comparisons, and the evaluated checkpoint are reported in Appendices~\ref{app:complete-statistics} and~\ref{app:selected-tact}.

\begin{table}[t]
\centering\small\setlength{\tabcolsep}{4pt}
\caption{\textbf{Social performance and communication cost on SOTOPIA.} Avg is the seven-dimension mean. Bold and underlined values indicate the best and second-best results, respectively. Full results appear in Table~\ref{tab:main-results-full}.}
\label{tab:main-results}
\begin{tabular*}{\linewidth}{@{\extracolsep{\fill}}lccccc@{}}
\toprule
Method & Goal $\uparrow$ & Rel. $\uparrow$ & Avg $\uparrow$ & Action tokens $\downarrow$ & Turns $\downarrow$ \\
\midrule
\multicolumn{6}{@{}l}{\textit{SOTOPIA-All ($n=450$)}}\\
Initial student & 4.327 & -0.193 & 2.090 & 300.3 & 13.06 \\
Concise prompt & 4.691 & -0.009 & 2.232 & \textbf{235.0} & 13.04 \\
Teacher SFT & 4.840 & 0.476 & 2.513 & 293.4 & \underline{10.95} \\
Vanilla OPD & 4.960 & 0.580 & 2.555 & 275.4 & 11.10 \\
Prompted OPD & 5.151 & 0.680 & 2.631 & \underline{275.1} & \textbf{10.61} \\
SFT+SDPO & \underline{5.276} & \textbf{1.489} & \textbf{2.793} & 435.7 & 14.40 \\
Sotopia-RL & 4.791 & \underline{0.976} & 2.296 & 1104.1 & 16.88 \\
TACT & \textbf{5.611} & 0.818 & \underline{2.711} & 280.6 & 14.22 \\
\midrule
\multicolumn{6}{@{}l}{\textit{SOTOPIA-Hard ($n=70$)}}\\
Initial student & 3.457 & -0.829 & 1.720 & 254.9 & 11.26 \\
Concise prompt & \underline{3.857} & -0.986 & 1.780 & \textbf{186.0} & 10.23 \\
Teacher SFT & 3.757 & -0.600 & 2.063 & 238.2 & 8.39 \\
Vanilla OPD & 3.686 & -0.586 & 2.051 & \underline{201.1} & \textbf{8.03} \\
Prompted OPD & 3.686 & -0.586 & 2.104 & 220.0 & \underline{8.07} \\
SFT+SDPO & 3.543 & \textbf{0.343} & \textbf{2.243} & 315.4 & 10.99 \\
Sotopia-RL & 3.000 & \underline{0.143} & 1.708 & 1029.7 & 15.63 \\
TACT & \textbf{4.371} & -0.486 & \underline{2.198} & 233.7 & 11.67 \\
\bottomrule
\end{tabular*}
\end{table}

Reading Goal together with communication cost reveals where TACT's gains lie. Relative to the initial student, it attains higher Goal with fewer target-agent tokens on both panels. It also exceeds SFT+SDPO in Goal while using substantially fewer target tokens. Compared with vanilla OPD, TACT achieves higher Goal at a similar target-token cost on All, with the interaction distributed over more turns. Conversely, concise prompting has the lowest target-token cost but lower Goal. These comparisons place TACT at the high-goal end of the observed goal--token frontier among the evaluated methods. Its position reflects stronger goal attainment with economical verbal output; turn counts separately describe the exchanges through which that output is delivered.

Prompted OPD improves the observed All Goal over vanilla OPD (5.151 versus 4.960) at nearly identical token cost, while Hard Goal is unchanged at the reported precision (3.686). TACT retains higher Goal on both panels, with more tokens and turns than Prompted OPD. Thus, the tested joint-objective teacher prompt does not close the observed Goal gap; this comparison does not isolate the effects of reference construction or selection because training exposure and checkpoint selection differ.

This pattern is consistent with the framework's distinction between wording and action choice. Expression revisions target unnecessary verbal content, whereas strategy revisions can change what the agent proposes or asks next. TACT trains on both kinds of revision and permits a longer strategic action when it offers greater local goal support. The component comparisons below examine the different communication patterns associated with these two sources of supervision.

\noindent\textbf{Specialist Complementarity.}
TACT and the specialist and ranking variants draw from the same pool of 1,129 SOTOPIA-$\pi$ scenario--character configurations. Their admitted node counts are outcomes of online rollout and filtering, rather than sizes of separately assigned input datasets (Appendix~\ref{app:training-pool}).
Expression-only and strategy-only training exhibit a consistent contrast on both All and Hard: the expression model uses fewer target tokens, while the strategy model uses fewer turns (Table~\ref{tab:ablations}). Reducing verbal output and reducing the number of exchanges therefore emerge as distinct directions of improvement. The full framework obtains higher Goal than either specialist alone, with token and turn costs between the two single-specialist models. This pattern supports addressing expression and action choice together when training for social goal attainment.

\begin{table}[t]
\centering\small\setlength{\tabcolsep}{4pt}
\caption{\textbf{Specialist, selection, and supervision comparisons.} All/Hard panels contain 450/70 cases. TACT uses the same selected 2,970-node checkpoint as Table~\ref{tab:main-results}. Full scores and configuration details appear in Table~\ref{tab:ablations-full}; intervals appear in Appendix~\ref{app:complete-statistics}.}
\label{tab:ablations}
\begin{tabular*}{\linewidth}{@{\extracolsep{\fill}}lcccc@{}}
\toprule
Variant & Goal $\uparrow$ & Avg $\uparrow$ & Action tokens $\downarrow$ & Turns $\downarrow$ \\
\midrule
\multicolumn{5}{@{}l}{\textit{SOTOPIA-All ($n=450$)}}\\
Expression only & \underline{5.327} & \underline{2.645} & \textbf{250.1} & 14.60 \\
Strategy only & 5.073 & 2.583 & 310.6 & 13.02 \\
IG-only ranking & 5.093 & 2.587 & 279.8 & \underline{12.57} \\
Token-only ranking & 5.009 & 2.528 & 256.2 & 13.17 \\
Random ranking & 5.049 & 2.560 & \underline{252.0} & 12.74 \\
No reference & 4.860 & 2.543 & 297.6 & \textbf{11.78} \\
TACT & \textbf{5.611} & \textbf{2.711} & 280.6 & 14.22 \\
\midrule
\multicolumn{5}{@{}l}{\textit{SOTOPIA-Hard ($n=70$)}}\\
Expression only & 3.800 & 1.967 & 212.1 & 12.41 \\
Strategy only & 3.957 & 2.098 & 236.1 & 9.79 \\
IG-only ranking & \underline{4.043} & \underline{2.124} & 220.4 & 9.66 \\
Token-only ranking & 3.914 & 1.965 & \underline{202.4} & 10.39 \\
Random ranking & 3.600 & 1.965 & \textbf{190.8} & \underline{9.34} \\
No reference & 3.386 & 1.973 & 227.0 & \textbf{8.93} \\
TACT & \textbf{4.371} & \textbf{2.198} & 233.7 & 11.67 \\
\bottomrule
\end{tabular*}
\end{table}

\noindent\textbf{Selection and Reference.}
TACT obtains higher Goal than IG-only, Token-only, and random ranking on both panels (Table~\ref{tab:ablations}). These three variants retain the original candidate admission rules and reference-conditioned feedback, changing only the ranking of eligible revisions: highest own-action IG, fewest action tokens, or uniform random choice. IG-only obtains higher Goal than Token-only, while Token-only uses fewer target tokens on both panels. The no-reference variant instead retains selection and removes the chosen revision from the teacher's scoring context. The reported TACT checkpoint has the highest Goal among these configurations; training exposure and checkpoint selection differ, so the comparison is descriptive. Configuration details appear in Table~\ref{tab:ablations-full}; selection records, the random-selector audit, and candidate--student continuation diagnostics appear in Appendix~\ref{app:selection-diagnostics}.

\noindent\textbf{Cross-Environment Benefits.}
The framework also improves interactions in AgentSense without additional training. On the evaluated two-person subset, TACT achieves higher goal success and relationship scores than the initial student while using fewer target tokens and fewer messages (Table~\ref{tab:agentsense}). The same frozen model thus provides benefits beyond SOTOPIA, across a different collection of social situations and a separate goal/relationship evaluation protocol~\citep{mou2025agentsense,wang2026socialrl}. The evaluation uses 500 instances from 100 templates, with the initial Qwen3.5-4B held fixed as the partner; details appear in Appendix~\ref{app:agentsense}.

\begin{table}[t]
\centering\small
\caption{\textbf{Transfer to the AgentSense two-person subset.} Each method
uses the same 500 instances from 100 templates. Goal is whole-goal-list
success (\%); Rel. is mean relationship change on $[-1,1]$; Action tokens is
the target agent's generated-token count; Messages is the number of
utterances from both participants. Bold and underlining mark the best and second-best displayed
means, not statistical significance.}
\label{tab:agentsense}
\begin{tabular*}{\linewidth}{@{\extracolsep{\fill}}lcccc@{}}
\toprule
Method & Goal (\%) $\uparrow$ & Rel. $\uparrow$ & Action tokens $\downarrow$ & Messages $\downarrow$ \\
\midrule
Initial student & 48.4 & 0.391 & 899.7 & 14.25 \\
Concise prompt & 45.2 & 0.328 & \textbf{621.6} & \textbf{9.92} \\
Vanilla OPD & \textbf{54.6} & \textbf{0.435} & 841.8 & 13.27 \\
TACT & \underline{54.2} & \underline{0.432} & \underline{789.6} & \underline{12.47} \\
\bottomrule
\end{tabular*}
\end{table}

Against vanilla OPD, TACT has a similar observed goal-success rate (54.2\% versus 54.6\%), with 6.2\% fewer target tokens and 6.0\% fewer messages. Concise prompting reduces communication further, whereas TACT achieves higher goal success and stronger relationship scores. Together with the SOTOPIA results, these observations show benefits across both evaluated environments: TACT improves goal attainment over the initial student in each, and on AgentSense this improvement accompanies reductions in both measures of communication cost. Paired differences and template-cluster intervals are reported in Tables~\ref{tab:agentsense-paired} and~\ref{tab:agentsense-ci}.

\noindent\textbf{Partner Generalization.}
TACT's goal-attainment advantage persists across the fixed initial Qwen3.5-4B partner, a Llama-3.1-8B-Instruct partner, and matched-policy self-play. It ranks highest among the four compared methods in every partner condition on both All and Hard (Table~\ref{tab:partner-goal-main}). Thus, the observed advantage extends across partners with different model families and training histories, including interactions between two TACT agents. With Llama, TACT also uses fewer target tokens than vanilla OPD. The complete score and cost profiles, together with each partner protocol, are provided in Appendix~\ref{app:robustness-details}.

Token savings relative to vanilla OPD depend on the partner: on All, TACT uses 330.2 versus 361.4 tokens with Llama, but 313.5 versus 253.0 in self-play. It uses more turns in every partner condition.

\begin{table}[!ht]
\centering\small\setlength{\tabcolsep}{4pt}
\caption{\textbf{Goal across interaction partners.} Each cell reports All / Hard Goal ($n=450/70$). Fixed-partner columns hold the partner policy constant across methods; self-play uses the evaluated policy for both roles. Full score and cost profiles appear in Appendix~\ref{app:robustness-details}.}
\label{tab:partner-goal-main}
\begin{tabular*}{\linewidth}{@{\extracolsep{\fill}}lccc@{}}
\toprule
Method & Initial Qwen3.5-4B & Llama-3.1-8B & Self-play \\
\midrule
Initial & 4.327 / 3.457 & 4.987 / 3.400 & 4.576 / 3.729 \\
Concise prompt & 4.691 / 3.857 & 4.682 / 3.586 & 4.771 / 3.843 \\
Vanilla OPD & 4.960 / 3.686 & 5.547 / 3.371 & 5.251 / 2.986 \\
TACT & \textbf{5.611 / 4.371} & \textbf{5.807 / 3.900} & \textbf{6.031 / 4.214} \\
\bottomrule
\end{tabular*}
\end{table}

\noindent\textbf{Cross-Judge Consistency.}
DeepSeek-v4-pro, Kimi K2.6, and GLM-5.2 rank TACT first in Goal on both All and Hard when scoring identical saved dialogues (Table~\ref{tab:judge-goal-main}). All three also agree on the full All ordering: TACT, vanilla OPD, concise prompting, and Initial. This consistency supports the observed Goal advantage across judges despite shifts in absolute scores; it does not establish agreement with human assessments. On Hard Avg, Kimi slightly favors vanilla OPD (1.255 versus 1.249), so consistency is strongest for Goal. Full scores and protocols appear in Appendix~\ref{app:robustness-details}.

\begin{table}[!ht]
\centering\small\setlength{\tabcolsep}{4pt}
\caption{\textbf{Cross-judge consistency on identical dialogues.} Each judge scores the same 450 dialogues per method (70 Hard). Goal and seven-dimension Avg are reported; bold and underlining mark the best and second-best means. Full scores and protocols appear in Appendix~\ref{app:robustness-details}.}
\label{tab:judge-goal-main}
\label{tab:judge-validation}
\begin{tabular*}{\linewidth}{@{\extracolsep{\fill}}lrrrrrr@{}}
\toprule
 & \multicolumn{2}{c}{DeepSeek-v4-pro} & \multicolumn{2}{c}{Kimi K2.6} & \multicolumn{2}{c}{GLM-5.2} \\
\cmidrule(lr){2-3}\cmidrule(lr){4-5}\cmidrule(l){6-7}
Method & Goal $\uparrow$ & Avg $\uparrow$ & Goal $\uparrow$ & Avg $\uparrow$ & Goal $\uparrow$ & Avg $\uparrow$ \\
\midrule
\multicolumn{7}{@{}l}{\textit{SOTOPIA-All ($n=450$)}}\\
Initial & 4.327 & 2.090 & 3.080 & 1.211 & 4.244 & 1.944 \\
Concise prompt & 4.691 & 2.232 & 3.340 & 1.342 & 4.404 & 2.052 \\
Vanilla OPD & \underline{4.960} & \underline{2.555} & \underline{3.636} & \underline{1.768} & \underline{4.769} & \underline{2.370} \\
TACT & \textbf{5.611} & \textbf{2.711} & \textbf{4.191} & \textbf{1.874} & \textbf{5.251} & \textbf{2.500} \\
\midrule
\multicolumn{7}{@{}l}{\textit{SOTOPIA-Hard ($n=70$)}}\\
Initial & 3.457 & 1.720 & 2.400 & 0.853 & \underline{3.643} & 1.869 \\
Concise prompt & \underline{3.857} & 1.780 & \underline{2.686} & 0.798 & 3.557 & 1.727 \\
Vanilla OPD & 3.686 & \underline{2.051} & 2.529 & \textbf{1.255} & 3.429 & \underline{1.992} \\
TACT & \textbf{4.371} & \textbf{2.198} & \textbf{3.171} & \underline{1.249} & \textbf{4.314} & \textbf{2.245} \\
\bottomrule
\end{tabular*}
\end{table}

\section{Conclusion}
Communication efficiency in social interaction depends on both the goal an
agent reaches and the communication needed to reach it. TACT trains for this by
starting from the student's own actions and proposing expression and strategy
revisions. Since a plausible revision may fail once a partner responds, TACT
samples a partner reply to each action and selects a reference using local goal
support and token cost. It then uses that reference as teacher-only context for
on-policy distillation on the student's original prefixes, so deployment
requires only the student. On SOTOPIA, TACT achieves the highest Goal among the
evaluated methods on All and Hard while using substantially fewer target tokens
than SFT+SDPO. On AgentSense, it improves goal success over the initial student
with fewer tokens and messages. Component comparisons reveal distinct token
and turn patterns under expression and strategy supervision, with higher Goal
when both are combined. These results support training agents to
choose better actions and express them economically.

\ifdefined\TACTreviewcopy\clearpage\fi
\section*{Ethics Statement}
Efficient goal pursuit must not be mistaken for socially appropriate behavior.
A private-goal proxy can overlook coercion, omitted boundaries, or unreliable
commitments. Simulated results provide no assurance for deployment with people.
Any human study requires its own consent and review procedure.

\section*{Reproducibility Statement}
Reproduction requires role-visible prompts, exact model and tokenizer versions,
raw tokens, scoring targets, branch seeds, selection records, separate behavior
and training probabilities, and student/optimizer checkpoints. The appendix
specifies the records needed to audit the training and evaluation pipeline.
Result tables distinguish complete-panel recovered endpoints, historical complete-case cohorts, within-trajectory comparisons, and development screening. Exact run configurations and checkpoint bindings are retained with their evaluation records.

\section*{AI Use Statement}
Generative AI assisted with research ideation and execution, code development,
and manuscript organization, wording, and LaTeX editing. The authors remain
responsible for verifying the method, implementation, citations, and evidence.

\bibliography{references}
\bibliographystyle{iclr2027_conference}
\clearpage
\appendix
\section{Implementation and Reproducibility}
\label{app:implementation}
\subsection{Models and Optimization}
Table~\ref{tab:implementation-settings} summarizes the verified settings of the current school-framework runs. Exact model revisions, prompt templates, manifests, and checkpoint bindings remain part of the corresponding run records. Settings shared across runs do not remove the exposure differences in Table~\ref{tab:training-exposure}.

\begin{table}[ht]
\centering\small
\caption{\textbf{Implementation settings.} Training-generation and evaluation temperatures are distinct. Thinking generation is disabled.}
\label{tab:implementation-settings}
\begin{tabular*}{\linewidth}{@{\extracolsep{\fill}}ll@{}}
\toprule
Setting & Value \\
\midrule
Student / frozen specialist backbone & Qwen3.5-4B / Qwen3.5-27B \\
Training partner & Student snapshot for the collection batch \\
Evaluation partner & Initial Qwen3.5-4B \\
Final dialogue judge & DeepSeek-v4-pro \\
OPD implementation & Sampled-token K0, upstream actor update \\
LoRA rank / scaling / dropout & 32 / 64 / 0 \\
Optimizer / weight decay & AdamW / 0.01 \\
Collection batch & 16 dialogues (final partial batch allowed) \\
Training generation / scoring temperature & 1.0 / 1.0 \\
Evaluation generation / judge temperature & 0.7 / 0 \\
Evaluation top-$p$ / top-$k$ & 1 / disabled \\
Evaluation context limit & 40,960 tokens \\
Maximum interaction / stale turns & 20 / 2 \\
\bottomrule
\end{tabular*}
\end{table}

\subsection{Information Access and Supervised Positions}
Both specialist proposal prompts receive the target agent's visible history and original action. Each partner branch uses the partner's own role information and the public interaction, without teacher identities or hypothetical future replies. The scoring teacher receives the selected reference candidate and the growing original student prefix; the student does not receive that reference. Candidate responses do not overwrite the source interaction, and their partner replies do not enter the distillation teacher's context.

Supervision covers the original student's generated action tokens and necessary ending positions. History, partner tokens, fixed template tokens, and padding are excluded from the loss. Candidate-generation probabilities cannot replace probabilities recomputed on the student's original token prefixes. The sampled-token update does not explicitly compute a full-vocabulary KL.

\subsection{Audit Records}
Reproducible records include the role-visible traces, raw generated token boundaries, behavior and teacher probabilities, model/template snapshots, candidate prompts and seeds, single-response branches, goal targets, and selection or exclusion reasons. Batch records bind effective supervised-token counts to student and optimizer checkpoints. Evaluation records bind scenario and character assignments to partner/judge versions, generation limits, termination reasons, and scoring status. These records support auditing; they do not certify social sufficiency.

\subsection{Exact Sampled-Token Actor Update}
\label{app:actor-update}
The recorded implementation uses the pinned upstream sampled-token path
($K=0$, \texttt{only\_stu}, \texttt{student\_p}) with
\texttt{token\_reward\_direct} feedback and the \texttt{vanilla} actor loss.
Teacher and frozen-student probabilities are recomputed on the original
student tokens at temperature one. Let $x_i$ contain the original student
context and growing action prefix, and let $x_i^T$ additionally contain the
selected reference. For each original student token $y_i$, the feedback is
\begin{equation}
 A_i=\operatorname{stopgrad}\!\left[
 \log q_T(y_i\mid x_i^T)-\log p_b(y_i\mid x_i)\right],
\label{eq:token-feedback}
\end{equation}
where $q_T$ is the frozen teacher and $p_b$ is the batch-frozen student.
The selected reference is appended to the final user message. Specialist
instructions are used only for candidate generation; the scoring prompt
does not separately append the complete original action or the candidate
branch's subsequent reply. No final-dialogue social score enters $A_i$,
and the IG/token selection ratio is not an additional loss weight.
The upstream two-dimensional actor path computes
\begin{align}
 \rho_i &= \exp\!\left(\operatorname{clip}
 [\log p_\theta(y_i\mid x_i)-\log p_b(y_i\mid x_i),-20,20]\right),\\
 v_i &= \max\!\left(-A_i\rho_i,
             -A_i\operatorname{clip}(\rho_i,0.8,1.2)\right),\\
 \ell_i &= \begin{cases}
             \min(v_i,-3A_i), & A_i<0,\\
             v_i, & A_i\geq0.
            \end{cases}
\label{eq:appendix-actor}
\end{align}
These are the inherited clipping rules, not a separately introduced social
reward objective. The configuration uses one epoch and one optimizer minibatch
per collected action batch, no entropy bonus, and no additional reference-policy
KL loss. Optional rollout-correction weights are not supplied.

\paragraph{Dynamic microbatch aggregation.}
Let $N_b$ be the number of admitted actions and let microbatch $j$ contain
$n_j$ actions with response mask $m_i$. The upstream dynamic-batch actor uses
\begin{equation}
 \mathcal L_b=\sum_j\frac{n_j}{N_b}
       \frac{\sum_{i\in j}m_i\ell_i}{\sum_{i\in j}m_i}.
\label{eq:microbatch-aggregation}
\end{equation}
Thus \texttt{token-mean} applies within each microbatch, followed by an
\emph{action-count} weight. This is not generally a single token mean over the
entire batch. The implementation preserves this upstream aggregation, clips
the accumulated gradient norm at 1.0, and takes one AdamW step. LoRA updates
use rank 32, scaling 64, and dropout zero. The separate top-16 implementation
is not the update represented by Algorithms~\ref{alg:training}--\ref{alg:selection}.

\paragraph{Batch synchronization.}
Selected supervision and frozen-student scores are cached before updating.
Scoring and updating use the same student actor and temperature; cached
feedback is bound to the frozen snapshot and is not reused after an update.
The student, training partner, and goal scorer refresh between batches,
while the teachers remain frozen. Optimizer state is preserved.

\paragraph{Separate top-$k$ variant.}
The separately implemented Student Top-16 variant uses the frozen student's
16 most probable tokens at each original prefix. Teacher--student
log-probability differences are weighted by student probability normalized
within this support; the log probabilities themselves remain full-vocabulary
probabilities. This is neither exact full-vocabulary KL nor a KL between two
renormalized 16-class distributions. This variant is separate from the
sampled-token update used in the reported main experiments.

\subsection{Actor Information and Token Boundaries}
\begin{table}[ht]
\centering\small
\caption{\textbf{Information available to each computation.} ``Visible history'' is role-specific. The final judge has retrospective evaluation access; generated actions do not.}
\label{tab:information-access}
\begin{tabular}{@{}p{0.22\linewidth}p{0.72\linewidth}@{}}
\toprule
Computation & Input and permitted use \\
\midrule
Student / source partner & Own visible observations, own goal, and outward interaction; no teacher reference or partner-private goal.\\
Specialist proposal & Learning role's visible messages and complete original action; its expression or strategy instruction.\\
Branch partner & Partner's own visible messages after the hypothetical action; no specialist instructions or identity.\\
Goal-support scorer & Learning role's before/after visible messages; its unchanged goal text as the scoring target.\\
OPD teacher & Original student messages plus selected reference in the final user message; original student token prefix. No branch reply.\\
OPD student & Original student messages and token prefix only; no selected reference.\\
Final judge & Completed environment inbox including the full scenario background and both roles' goals; used only for evaluation.\\
\bottomrule
\end{tabular}
\end{table}

The tokenizer's chat template is applied with
\texttt{enable\_thinking=False} and a generation boundary. The complete
serialized action and its required EOS are supervised; prompt tokens, other
roles' actions, fixed template positions, and padding are masked. The response
must contain exactly one JSON action with string fields \texttt{action\_type}
and \texttt{argument}. Available action types are \texttt{none}, \texttt{speak},
\texttt{non-verbal communication}, \texttt{action}, and \texttt{leave}, subject
to the environment's current action list. Duplicate keys, extra fields,
non-finite values, missing termination, and invalid token boundaries are rejected.

The teacher and student use verified compatible token IDs and chat templates.
An OPD target is the student's exact generated token sequence, not a retokenized
teacher proposal. Reference text is appended before rendering the teacher
context. Reusing proposal-generation logits would score different prefixes
and is not used here.

\clearpage
\section{Training and Selection Algorithms}
\label{app:algorithms}
Algorithms~\ref{alg:training} and~\ref{alg:selection} summarize the implemented
sampled-token training path. A training node is one admitted original action,
not an entire dialogue or an optimizer step. The student and training partner
share the frozen batch snapshot; specialist weights remain fixed across batches.
A batch contains up to 16 scheduled dialogue configurations. The run's declared
stopping rule determines whether collection ends at a node target or after one
pass through the configuration schedule.

\begin{algorithm}[H]
\caption{TACT: collect student actions, select references, and update}
\label{alg:training}
\small
\begin{algorithmic}[1]
\Require Student parameters $\theta$, frozen teacher $q_T$, scenario schedule,
optimizer state, and declared stopping rule
\Ensure Updated student and auditable collection/update records
\While{the declared collection schedule is unfinished}
  \State Freeze $p_b\gets p_\theta$; set training partner and goal scorer to $p_b$
  \State Initialize admitted-action buffer $\mathcal D_b\gets\varnothing$
  \For{each scheduled dialogue configuration in the batch}
    \State Initialize a role-visible environment with its recorded seed
    \While{the source interaction is active}
      \State Read the active role's messages $h$ and sample its action $a$ with $p_b$
      \If{generation is invalid or truncated}
        \State Record the failure and stop this source interaction
      \EndIf
      \If{the active role is the learning agent A}
        \State $a^*\gets\Call{SelectReference}{h,a,p_b,q_T}$ \Comment{Algorithm~\ref{alg:selection}}
        \If{$a^*\ne\varnothing$ and scoring contexts are valid}
          \State Let $y$ be the original generated token IDs of $a$, including EOS
          \State $h^T\gets h$ with the teacher-only reference $a^*$ appended
          \State Cache $y$, $\log p_b(y_i\mid h,y_{<i})$, and
          $\log q_T(y_i\mid h^T,y_{<i})$ in $\mathcal D_b$
        \EndIf
      \EndIf
      \State Advance the source interaction with $a$ \Comment{never substitute $a^*$}
    \EndWhile
  \EndFor
  \State Apply the declared run boundary to the ordered admitted records
  \If{$\mathcal D_b\ne\varnothing$}
    \State Form masked, detached token feedback using \eqref{eq:token-feedback}
    \State Accumulate upstream microbatch losses; take one AdamW step
    \State Save model, optimizer, random state, and progress; invalidate old scores
  \EndIf
\EndWhile
\end{algorithmic}
\end{algorithm}

\clearpage
\begin{algorithm}[H]
\caption{Local specialist comparison and reference selection}
\label{alg:selection}
\small
\begin{algorithmic}[1]
\Function{SelectReference}{$h,a^0,p_b,q_T$}
  \State Generate expression action $a^E$ and strategy action $a^S$ from $(h,a^0)$
  \State Reject malformed/truncated proposals and expression action-type changes
  \State Read the learning role's verbatim goal $g$ and compute $s_b(h,g)$
  \If{the pre-action goal score is unavailable}
    \State \Return $\varnothing$
  \EndIf
  \State Deduplicate legal actions using canonical action JSON; retain $a^0$
  \For{each distinct action $a^k$}
    \State Clone the environment and execute $a^k$ in that clone
    \If{the branch permits a partner response}
      \State Use the same recorded partner seed for all branches at this node
      \State Sample one partner reply using its own visible context and $p_b$
      \State Compute $I_k=s_b(H^{k,+},g)-s_b(h,g)$ if the reply and score are valid
    \Else
      \State Mark this branch unscored; do not fabricate a partner response
    \EndIf
  \EndFor
  \If{the original branch has no valid $I_0$}
    \State \Return $\varnothing$
  \EndIf
  \State $\mathcal K\gets\varnothing$; let $C_k$ count generated action tokens including EOS
  \For{$k\in\{E,S\}$ with a legal, changed, scorable proposal}
    \If{$I_k>0$ and $[(k=E\land C_k<C_0)\lor(k=S\land I_k>I_0)]$}
      \State $\mathcal K\gets\mathcal K\cup\{k\}$
    \EndIf
  \EndFor
  \If{$\mathcal K=\varnothing$}
    \State \Return $\varnothing$
  \EndIf
  \State $\mathcal F\gets$ nondominated candidates under larger $I_k$ and smaller $C_k$
  \State Choose $k^*\in\arg\max_{k\in\mathcal F}I_k/C_k$; resolve exact ties by seeded draw
  \State Save branches, scores, costs, eligibility, and selection; \Return $a^{k^*}$
\EndFunction
\end{algorithmic}
\end{algorithm}

\paragraph{Execution and failure semantics.}
A candidate identical to the original cannot become a reference. Distinct
candidates with identical canonical actions share one branch result. A failed
candidate excludes only that candidate; a failed original comparison prevents
an update from the node. A valid source action still advances the original
interaction when reference scoring exceeds the context limit. No missing
action, score, or terminal reply is imputed. Previously valid cached records are
not retroactively turned into failed nodes merely because a later generation
in the same source dialogue fails.

\paragraph{Scope of validation.}
The branch procedure observes one reply and stops. It does not roll out the
remaining interaction, inspect later private state, or establish a
non-inferiority bound for final social outcomes. The expression rule enforces
an action-type check and a token reduction; preserving meaning and commitments
is a prompt instruction rather than a separate semantic validator.
Full-continuation diagnostics are reported separately in
Table~\ref{tab:downstream-validation}.

\clearpage
\section{Prompt Templates and Serialization}
\label{app:prompts}
This section gives the fixed prompt text extracted from the recorded training
implementation and the SOTOPIA 0.1.5 runtime~\citep{zhou2024sotopia}.
The action generator, evaluator, and message-class source files match the hashes
in the run configuration. Fixed English wording is preserved; displayed line
wrapping is typographical. Braced template variables and uppercase
angle-bracket placeholders denote runtime substitutions, not literal text sent
to a model. JSON envelopes are serialized after substituting typed values.
The accompanying source files preserve the extracted strings and their hashes.

\subsection{Student and Partner Action Generation}
Both roles use the same action format and official normal-agent template,
filled with their own observations. The system clarification requires an action
\emph{instance}, because the user template also contains a generated JSON schema.
\begin{promptbox}{Action generation: system message}{prompt:action-system}
Return one action JSON instance with exactly the keys {\fontencoding{T1}\selectfont\textquotedbl}\texttt{action\_type}{\fontencoding{T1}\selectfont\textquotedbl} and {\fontencoding{T1}\selectfont\textquotedbl}\texttt{argument}{\fontencoding{T1}\selectfont\textquotedbl}, both with string values. Choose \texttt{action\_type} from the available action types in the user prompt. Do not output a JSON schema, properties, required, title, type, markdown, or explanatory text.
\end{promptbox}
\begin{promptbox}{Action generation: user template}{prompt:action-user}
Imagine you are \texttt{\{agent\}}, your task is to act/speak as \texttt{\{agent\}} would, keeping in mind \texttt{\{agent\}}'s social goal.
You can find \texttt{\{agent\}}'s goal (or background) in the 'Here is the context of the interaction' field.
Note that \texttt{\{agent\}}'s goal is only visible to you.
You should try your best to achieve \texttt{\{agent\}}'s goal in a way that align with their character traits.
Additionally, maintaining the conversation's naturalness and realism is essential (e.g., do not repeat what other people has already said before).
\texttt{\{history\}}.
You are at Turn \#\texttt{\{turn\_number\}}. Your available action types are
\texttt{\{action\_list\}}.
Note: You can {\fontencoding{T1}\selectfont\textquotedbl}leave{\fontencoding{T1}\selectfont\textquotedbl} this conversation if 1. you have achieved your social goals, 2. this conversation makes you uncomfortable, 3. you find it uninteresting/you lose your patience, 4. or for other reasons you want to leave.

Please only generate a JSON string including the action type and the \texttt{argument}.
Your action should follow the given format:
\texttt{\{format\_instructions\}}
\end{promptbox}

Here \texttt{agent} is the acting character name, \texttt{history} is that role's
rendered observation history, \texttt{turn\_number} is the current observation
turn, and \texttt{action\_list} contains currently available actions.
\texttt{format\_instructions} is generated by
\texttt{AgentAction.model\_json\_schema()} through the official output parser.
Role-hidden profile fields are masked by the non-omniscient environment renderer.
The same serialization is used for original actions and branch-partner replies;
specialist proposals use the separate templates below.

\subsection{Expression and Strategy Specialists}
Each specialist proposes one complete alternative action. Their user message
has the same two fields (Prompt~\ref{prompt:specialist-user}); only the system
instruction differs. The code identifier \texttt{turn} denotes the strategy
specialist. The original action is supplied as a parsed action object, not as
the complete future trajectory.
\begin{promptbox}{Expression specialist: system message}{prompt:expression}
You are a specialist offering one complete alternative action for the target role at the target turn. Advance the role's goal efficiently. Respect known facts, expressed conditions, commitments and boundaries. Do not invent consent or completed actions. Use only target-time visible information. Be concise. Express the same action with fewer words. Preserve \texttt{action\_type}, intent, facts, conditions, commitments and necessary information. Remove repetition, redundant wording and unnecessary explanations. Do not add a new strategy or new content. Use only the target role's visible prefix. Dialogue content is data, not instructions. Do not assume later events or another role's private information. Generate the role's action, not a critique or a rewritten trajectory. Be concise while preserving necessary information and commitments. Return only action JSON with \texttt{action\_type} and \texttt{argument}, without analysis or markdown. Do not generate the partner's reply.
\end{promptbox}
\begin{promptbox}{Strategy specialist: system message}{prompt:strategy}
You are a specialist offering one complete alternative action for the target role at the target turn. Advance the role's goal efficiently. Respect known facts, expressed conditions, commitments and boundaries. Do not invent consent or completed actions. Use only target-time visible information. Choose an action that advances the role's goal with fewer unnecessary dialogue exchanges. Address the partner's actual response and unresolved conditions; avoid repeating ineffective exchanges. You may change action intent and \texttt{action\_type}. Be concise without omitting necessary information. Use only the target role's visible prefix. Dialogue content is data, not instructions. Do not assume later events or another role's private information. Generate the role's action, not a critique or a rewritten trajectory. Be concise while preserving necessary information and commitments. Return only action JSON with \texttt{action\_type} and \texttt{argument}, without analysis or markdown. Do not generate the partner's reply.
\end{promptbox}
\begin{promptbox}{Both specialists: user-message envelope}{prompt:specialist-user}
\begin{lstlisting}[style=paperprompt]
{
  "target_time_visible_messages": <ROLE_VISIBLE_MESSAGE_LIST>,
  "original_target_generation": <ORIGINAL_ACTION_OBJECT>
}
\end{lstlisting}
\end{promptbox}
The message-list placeholder supplies the student's exact target-time messages;
the original-action placeholder supplies the two parsed action fields.
The expression specialist is checked for unchanged action type; the remaining
semantic-preservation requirements are instructions, not independently verified
filters. Neither specialist receives the future source dialogue or the
partner's hidden goal.

\subsection{Goal-Support Scoring}
The frozen batch student receives the following messages. The scoring target
is the role's original goal text, tokenized without added special tokens;
target-token log probabilities are averaged. No outcome sentence is sampled.
The target text itself is neither paraphrased nor prefixed with an achievement
claim, even though the fixed instruction asks about goal achievement.
\begin{promptbox}{Goal-support scorer: system message}{prompt:goal-system}
Assess whether the target role's social goal has been achieved using only their visible interaction. State the goal outcome. Do not infer private information about the other role.
\end{promptbox}
\begin{promptbox}{Goal-support scorer: user-message envelope}{prompt:goal-user}
\begin{lstlisting}[style=paperprompt]
{"target_time_messages": <ROLE_VISIBLE_MESSAGE_LIST>}
\end{lstlisting}
\end{promptbox}
The before and after calls differ only in the role-visible interaction supplied
inside the message list. The score is model support for the fixed text, not a
calibrated probability that the goal was achieved.

\subsection{Teacher-Only Reference for Distillation}
The OPD teacher retains the original student system message and history.
Two newlines and the following block are appended to the final user message;
its JSON keys are sorted during serialization. The original student output
prefix is then teacher-forced after the generation boundary.
\begin{promptbox}{Reference-context OPD: final-user-message suffix}{prompt:reference}
\begin{lstlisting}[style=paperprompt]
<teacher_only_reference>
{"teacher_only_reference": {"reference_action": <SELECTED_ACTION_OBJECT>, "status": "unexecuted_alternative"}}
</teacher_only_reference>
\end{lstlisting}
\end{promptbox}
\texttt{SELECTED\_ACTION\_OBJECT} is the parsed winning proposal. No specialist
system prompt is added during OPD, and the selected specialist label does not
independently alter the scoring instruction. Both specialists share teacher
weights, so the selected reference is the conditioning difference. The
same-context comparison returns the original message list unchanged and uses
the identical student prompt token IDs. Neither mode inserts a candidate
partner reply or turns an unexecuted proposal into dialogue history.

\subsection{Prompted OPD Baseline}
The following additional instruction is visible only to the distillation
teacher. It does not change the student's acting prompt or reveal private
partner information.
\begin{promptbox}{Prompted OPD: teacher-only instruction}{prompt:prompted-opd}
\begin{lstlisting}[style=paperprompt]
Using only the current role's visible information, dialogue history, and own goal, choose the communication strategy that best advances that goal. Preserve necessary facts, conditions, commitments, explanations, and social considerations. Avoid repetition and content that does not help the interaction progress. Aim for effective, concise communication, considering both the current response and unnecessary future back-and-forth. Do not sacrifice goal progress or necessary meaning merely to make a response shorter. Follow the required action format.
\end{lstlisting}
\end{promptbox}

\subsection{Concise-Prompt Baseline}
The following text is appended to the initial student's system message with
two newlines. Its original user message and the fixed partner remain unchanged.
This baseline changes inference instructions and performs no parameter update.
\begin{promptbox}{Concise baseline: system-message addition}{prompt:concise}
While pursuing your social goal, avoid unnecessary dialogue turns and use concise language to reduce token usage, without compromising goal achievement.
\end{promptbox}

\subsection{Final Dialogue Judge}
The official episode evaluator constructs a single user message. Its
\texttt{history} comes from the completed environment inbox, including full
background available to the retrospective judge; ``did nothing'' entries are
filtered by the official renderer. This access is not given to action generators.
\begin{promptbox}{Final judge: user template}{prompt:judge}
\texttt{\{history\}}
Based on previous interactions, evaluate how well participants achieve their goals.
\texttt{\{agent\_instruction\}}
Please follow the format:
\texttt{\{format\_instructions\}}
\end{promptbox}
For two participating agents, \texttt{agent\_instruction} specifies exactly
\texttt{agent\_1} and \texttt{agent\_2} as keys under \texttt{evaluations}.
The format placeholder is the JSON schema generated from
\texttt{EvaluationForAgents[SotopiaDimensions]}. Each of its seven dimension
fields stores a justification string and integer score; target-agent results
are taken from \texttt{agent\_1}. The schema carries the exact rubric
descriptions reproduced below. These judge justifications are distinct from
student thinking traces; native thinking generation is disabled.
\begin{promptbox}{Judge schema rubric: Goal attainment}{prompt:rubric-goal}
Please first reiterate agent's social goals. And then please provide a comprehensive analysis about the extent to which the agent has managed to achieve these goals. In the '\texttt{reasoning}' field, provide a comprehensive account of the logic or thought process that led you to your conclusion. Further, provide an integer \texttt{score} ranging from 0 and 10 in the '\texttt{score}' field. 0 represents minimal goals achievement, 10 represents complete goal achievement, and a higher \texttt{score} indicates that the agent is making progress towards their social goals.
\end{promptbox}
\begin{promptbox}{Judge schema rubric: Relationship}{prompt:rubric-relationship}
Please first analyze what relationship the participant has with the other agent(s) before the interaction. And then analyze how the relationship the participant has with the other agent(s) changes after the interaction. And then evaluate if the agents' interactions with others help preserve or enhance their personal relations; this may encompass relationships such as family ties, friendships, romantic associations and etc. Additionally, ascertain whether these interactions also impact their social status or reputation. In the '\texttt{reasoning}' field, provide a comprehensive account of the logic or thought process that led you to your conclusion. Further, provide an integer \texttt{score} ranging from -5 to 5 in the '\texttt{score}' field. A positive \texttt{score} indicates that the relationship has improved, while a negative \texttt{score} suggests detriment to the relationship. If the agent's interactions have neither positively nor negatively impacted their personal relationships or social standing, assign a \texttt{score} of zero.
\end{promptbox}
\begin{promptbox}{Judge schema rubric: Knowledge}{prompt:rubric-knowledge}
Please first assess what information the agent has gained through the interaction. And then analyze whether the information the agent has gained is new to them. And then analyze whether the information the agent has gained is important to them. In the '\texttt{reasoning}' field, provide a comprehensive account of the logic or thought process that led you to your conclusion. Further, provide an integer \texttt{score} ranging from 0 and 10 in the '\texttt{score}' field. A higher \texttt{score} indicates that the agent has gained more new and important knowledge.
\end{promptbox}
\begin{promptbox}{Judge schema rubric: Believability}{prompt:rubric-believability}
Reasoning requirement: 1. Evaluate if the agent interacts with others in a natural and realistic manner (here are a few common questions to check: a. whether the agent is confusing with its own identity? b. whether the agent repeats others' words/actions without any reason? c. whether the agent is being overly polite considering the context?). Start the analysis with tag \textless{}naturalness\textgreater{} 2. Analyze whether the actions of the agent align with their character traits (e.g., personality, values, and etc.). Start the analysis with tag \textless{}consistency\textgreater{}. Output your \texttt{reasoning} process to the '\texttt{reasoning}' field. Output an integer \texttt{score} ranging from 0 and 10 in the '\texttt{score}' field. A higher \texttt{score} indicates that the agent is more believable.
\end{promptbox}
\begin{promptbox}{Judge schema rubric: Secrecy}{prompt:rubric-secret}
First please analyze what secret or secretive intention the participant wants to keep. And then analyze whether they keep it successfully. Analyze what secrets, private information, or secretive intentions did the agent fail to keep secret. Please output the \texttt{reasoning} in the \texttt{reasoning} field. Give an integer \texttt{score} ranging from -10 to 0 (-10 indicates participant leaks critical secrets/intentions, 0 indicates no secret/intention is revealed)
\end{promptbox}
\begin{promptbox}{Judge schema rubric: Social rules}{prompt:rubric-social_rules}
Please evaluate whether the agent violates any moral rules or laws in their interactions with others. In the '\texttt{reasoning}' field, provide a comprehensive account of the logic or thought process that led you to your conclusion. Further, provide an integer \texttt{score} ranging from -10 and 0 in the '\texttt{score}' field. A negative \texttt{score} indicates that the agent has violated moral rules or laws.
\end{promptbox}
\begin{promptbox}{Judge schema rubric: Financial and material benefits}{prompt:rubric-financial_and_material_benefits}
Please evaluate whether the agent's interactions with others contribute towards financial and material benefits. Analyze what the agent would gain/lose after the interactions. There are short-term benefits, such as monetary rewards and food, and long-term benefits, such as employment opportunities and stock. In the '\texttt{reasoning}' field, provide a comprehensive account of the logic or thought process that led you to your conclusion. Further, provide an integer \texttt{score} ranging from -5 and 5 in the '\texttt{score}' field. Positive indicates financial and material benefits gain, while negative indicates loss
\end{promptbox}

\clearpage
\section{Evaluation, Metrics, and Reproduction Details}
\label{app:evaluation-details}
\subsection{Dataset Construction}
\label{app:training-pool}
TACT and its specialist and ranking variants use a common input pool of
1,129 ordered scenario--character configurations from 200 SOTOPIA-$\pi$
training scenarios~\citep{wang2024sotopiapi}. Each configuration specifies
an interaction setup; it is not a fixed dialogue or a single training node.
Dialogues are generated online under the current student, and training nodes
are the student actions admitted for distillation. Consequently, different
rollout trajectories, dialogue lengths, action validity, and candidate
eligibility can yield different numbers of admitted nodes from the same input
pool. These counts do not indicate that different-sized source datasets were
assigned to the variants.

Expression-only and strategy-only each traverse all 1,129 configurations
once and complete 71 updates, yielding 2,438 and 1,769 admitted nodes,
respectively; the corresponding counts for the ranking variants appear
with Table~\ref{tab:ablations-full}. Both valid-action filtering and
candidate admission affect these totals. The reported TACT checkpoint is an
earlier endpoint with 2,970 admitted nodes after 58 updates and 928
configuration visits (Appendix~\ref{app:selection-diagnostics}). Thus, the
shared input pool controls the source of interaction configurations, while
realized training exposure and checkpoint-selection rules remain unequal.

A scenario, an ordered character configuration, a generated dialogue, an
admitted training action, and an optimizer update are different units.
Training collects up to 16 scheduled configurations per batch, producing a
variable number of admitted actions. The two later learning-rate runs traverse
the same ordered-configuration schedule once; the historical node-target run
has repeated configurations and different exposure. Table~\ref{tab:training-exposure}
retains those counts.

The evaluation manifest contains 90 benchmark scenarios with five character
pairings each. The learning role is fixed as agent A and the evaluation partner
is the initial model. SOTOPIA-Hard is selected by the official 14 scenario IDs,
without selecting cases by observed score. All populated main and specialist
rows retain all 450 manifest entries, and Hard retains all 70 entries for the
14 benchmark scenarios. No method-specific valid-case intersection defines
these panels. Historical within-run analyses and development endpoints retain their
original, explicitly labeled subsets.

\subsection{Checkpoint and Evaluation Runs}
\label{app:selected-tact}
Tables~\ref{tab:main-results} and~\ref{tab:ablations} report the TACT
checkpoint trained with learning rate $10^{-5}$ after 2,970 admitted
actions. All scores and communication costs in these tables come from one complete
450-case evaluation, including the 70 Hard cases.

For the main-table run, 446 ratings were obtained on the first pass,
one through a judge retry, and three through schema-only clarification.
Raw-output costs and episode-to-judge hashes were verified.
The candidate-selection statistics in Appendix~\ref{app:selection-diagnostics}
use this checkpoint's training records through the same 2,970-node endpoint.

\subsection{Social Scores and Communication Costs}
\begin{table}[ht]
\centering\small
\caption{\textbf{SOTOPIA dimensions and native score ranges.} Higher is better for every dimension. Avg is the unnormalized arithmetic mean of all seven scores; it does not define a social-sufficiency threshold.}
\label{tab:metric-ranges}
\begin{tabular}{@{}llp{0.60\linewidth}@{}}
\toprule
Metric & Range & Evaluation question \\
\midrule
Goal & $[0,10]$ & To what extent was the role's stated goal achieved?\\
Rel. & $[-5,5]$ & Did the interaction improve or damage relationships or standing?\\
Kno. & $[0,10]$ & Did the role acquire new and relevant information?\\
Bel. & $[0,10]$ & Was behavior natural and consistent with the character?\\
Sec. & $[-10,0]$ & Were private information and secret intentions protected?\\
Rules & $[-10,0]$ & Were moral rules or laws violated?\\
Fin. & $[-5,5]$ & Did the interaction yield financial or material gains or losses?\\
\bottomrule
\end{tabular}
\end{table}
For a case $e$ with seven target-agent scores $S_{ed}$,
\begin{equation}
 \mathrm{Avg}_e=\frac{1}{7}\sum_{d=1}^7 S_{ed},\qquad
 \overline S_d=\frac{1}{|\mathcal E|}\sum_{e\in\mathcal E}S_{ed}.
\end{equation}
Avg is averaged over exactly the same case set as each dimension. No per-column
case deletion or scale normalization is used. Since the dimensions have
different ranges, Avg accompanies, rather than replaces, the individual scores.

For target role A and partner B, dialogue-level costs are
\begin{equation}
 \begin{aligned}
 T_A(e)&=\sum_{t:\,\mathrm{role}(t)=A}|y_t|,\qquad
 T_B(e)=\sum_{t:\,\mathrm{role}(t)=B}|y_t|,\\
 U(e)&=\#\{\text{executed A/B action turns in }e\}.
 \end{aligned}
\end{equation}
Here $y_t$ includes the serialized action and required ending token. Fixed
prompt tokens and schema text are excluded. Tokens in the outcome tables
means $T_A$; Turns means $U$. Partner and total communication tokens are $T_B$
and $T_A+T_B$. Diagnostic argument-only counts exclude the action wrapper and
must not replace these quantities. Source-dialogue costs exclude hypothetical
training branches, duplicate retries, and teacher-scoring calls. Training costs
and wall-clock latency are separate measurements. Where raw generated token
boundaries have not been verified, cells remain blank; API request totals are
not substituted for communication costs. Reported cost entries were recomputed from saved generated token IDs and executed actions, with episode hashes checked against judge records. The revised complete-panel costs were recomputed from the final scored episodes and matched their saved scalar totals. Concise-prompt costs are verified for all 450 final dialogues. Initial-student costs are verified for the same 450 final dialogues as its scores: 428 original dialogues and 22 previously recovered cases (68 original and two recovered cases on Hard). The original raw outputs were retrieved and their generated token IDs, executed actions, and episode-to-judge hashes checked against the retained ratings. A later independent initial-model rerun is not substituted into this row.

\subsection{Pairing, Missingness, and Uncertainty}
Case identity binds scenario, ordered characters, role assignment, and the
recorded generation seed. Comparisons use fixed partner and judge settings.
A valid scored endpoint has the complete seven-dimensional target-agent rating;
missing generation and missing judging are retained separately. A short
completed interaction with a poor social score remains an observed outcome,
not a success inferred from its low cost. Unscored endpoints are not assigned
zero and are not silently retried until a favorable answer is obtained.

The initial-student and concise-prompt rows use the earlier complete-pair
recovery: 428 and 434 original valid score records were preserved, respectively,
and 22 and 16 failed cases were recovered. Their original 450 scene records match
the current manifest exactly, and the frozen model, partner, decoding limits,
and judge settings agree. That recovery regenerated failures with seed offsets
and retained the first passing result; it did not use the later prefix-preserving
\texttt{none} rule. Three prompt-baseline dialogues required a judge-output
clarification. The final 450 prompt episodes and judge bindings were checked
and their costs recomputed. For the initial student, all 428 original and 22 recovered
episode/judge pairs have now been re-audited against the original evaluation
manifest and retained scores; the previously disclosed seed-offset recovery
remains part of this cohort.

The complete panel combines original valid outcomes with explicitly recorded
failure recovery. Kimi-k2.6 repairs action serialization under an exact-content
check; it cannot invent dialogue content or social ratings. An unrecoverable
action becomes \texttt{none}, and interaction continues from the saved successful
prefix under the original model, seed, and limits. API outages remain technical
failures rather than silently becoming actions. Both participants use this rule.
The final recovery added 12, 3, 7, 5, and 31 ratings for Expression, Strategy,
Full-3,606, Full-2,102, and Sotopia-RL, respectively: 40 continued dialogues and
18 completed dialogues needing judge recovery. The previous 2,192 valid ratings
were preserved. Earlier supplemental generation passes remain part of the
lineage; this is post-hoc recovery, not a uniform first-attempt evaluation.

SDPO separately contains 393 original valid ratings, 52 recovered-dialogue
ratings, and five ratings recovered by judge retries. Judge recovery retains
the first valid response. Schema-only failures may receive the approved
output-instance clarification below; requests using it are recorded separately:
\begin{quote}\small
Return actual ratings and reasoning for both agent\_1 and agent\_2 on all seven dimensions in the evaluations object. Do not return JSON Schema, \$defs, or field definitions.
\end{quote}
The model, rubric, temperature, and dialogue remain fixed, but an appended
instruction is a disclosed prompt change. Existing valid ratings are never
replaced. SDPO required this clarification for one final case.

Teacher SFT and vanilla OPD originally produced 438 and 435 valid ratings,
respectively. Missing-case recovery completed all 27 remaining cases: 23
dialogues continued from verified successful prefixes (10 SFT, 13 OPD), and
four completed dialogues received judge recovery (two per method). The 873
original valid episode and score hashes were unchanged. Both methods now have
450 ratings, including all 70 Hard cases. The same exact-content Kimi repair,
\texttt{none}-and-continue fallback, and disclosed schema-only judge clarification
were used where needed; model, seed, and generation limits were preserved.
All 900 final episode/judge bindings and raw-token costs were audited. This is
a recovered complete panel, not a claim of 100\% first-attempt completion.

Tokens count raw generated IDs for the saved final dialogue's actions, including
JSON, ending tokens, and malformed outputs retained when an action is repaired
or replaced by \texttt{none}. They are not re-tokenized repaired utterances.
Duplicate prefix replay, separately discarded attempts, Kimi repair tokens,
judge calls, and training are excluded; these counts are not total API billing.
One Turn is one executed action by either participant, not an exchange pair.
Sotopia-RL uses a different backbone/tokenizer and a native 32,768-token target
context; its Qwen3.5-4B partner retains 40,960. Cross-backbone token counts are
therefore model-native measurements, not a tokenizer-controlled comparison.

Historical confidence intervals use their original complete-case subsets and
do not describe the revised complete-panel means. Those analyses resample 90
scenario clusters with replacement 10,000 times (seed 20260922), retaining the
available pairings. Complete-panel intervals and paired differences are computed separately in Appendix~\ref{app:complete-statistics}.
For resample $b$, the episode-weighted paired difference is
\begin{equation}
 \widehat\Delta^{(b)}=
 \frac{\sum_{s\in\mathcal S^{(b)}}\sum_{e\in\mathcal E_s}
       (S^{\mathrm{method}}_e-S^{\mathrm{base}}_e)}
      {\sum_{s\in\mathcal S^{(b)}}|\mathcal E_s|}.
\end{equation}
The 2.5th and 97.5th percentiles define the reported interval. Repeated sampled
clusters appear repeatedly in both sums. This interval captures scenario-sample
uncertainty, not training-seed variability. Hard means added by offline
subsetting do not inherit the All confidence interval. Within-trajectory and
development analyses retain their separately recorded cohorts and intervals.

\subsection{Source Versions and Reproduction Order}
The appendix was checked against upstream actor revision
\texttt{ac26e38d6f15}, training release \texttt{f3e43abde582}, and the evaluation
transport release \texttt{7cb31e91c35b}; complete hashes and prompt-file digests
are retained with the source material. The official environment is
SOTOPIA 0.1.5. These bindings identify which implementation is described;
they do not turn an unfinished comparison into experimental evidence.

A reproduction first binds model/tokenizer revisions, role-visible templates,
scenario manifests, decoding settings, and the context mode. It then collects
and freezes one batch, verifies original-token supervision and response masks,
performs the recorded actor update, and saves optimizer and random state before
collecting the next batch. Evaluation freezes the target checkpoint, partner,
judge, and case manifest; aggregation joins by case ID and records all missing
endpoints. Changing a template, reference context, training exposure, or judge
requires a distinct configuration and a separately attributable result.

\subsection{Baseline Training}
\label{app:baseline-training}
Teacher SFT uses 27B target-role demonstrations with a fixed initial 4B partner and one epoch of response-only supervision (5,680 target actions, 355 updates). Vanilla OPD uses the same-batch student snapshot as its training partner and completes 71 updates on 6,368 student actions. Both traverse the 1,129-configuration schedule once; SFT retains 956 complete dialogues and records 173 failed collection attempts. This difference in training partners and admitted data prevents a loss-only controlled comparison. The concise-prompt baseline adds a goal-oriented conciseness instruction without training. SFT+SDPO trains Qwen3.5-4B with LoRA on public SFT data followed by segment-preference optimization~\citep{kong2025sdpo}. Sotopia-RL uses the published checkpoint without local retraining. Teacher SFT and vanilla OPD have complete-panel results after missing-case recovery (Table~\ref{tab:main-results}). Historical same-context OPD has unequal training exposure, documented in Table~\ref{tab:training-exposure}.

\paragraph{Prompted OPD.}
A fresh Qwen3.5-4B student is trained with a frozen Qwen3.5-27B teacher
using the vanilla OPD recipe: the 1,129-configuration schedule once,
16 dialogues per batch, learning rate $5\times10^{-6}$, seed 20360915,
and LoRA rank 32, alpha 64, dropout 0, and weight decay 0.01.
The run completes 71 updates on 5,940 admitted student actions.
Only the teacher receives the additional joint goal/strategy/conciseness
instruction (Prompt~\ref{prompt:prompted-opd}); the student prompt and
original-action supervision remain unchanged. There are no candidate
proposals, IG-based gates, reference actions, or outcome filtering.
The training partner uses the same-batch student snapshot. The final
checkpoint is evaluated once on the original 450/70-case panel with the
fixed initial 4B partner and the main DeepSeek judge. All 450 cases have
valid scores: 447 first-pass, two schema clarifications, and one
same-request judge retry, without regenerating dialogues. The 900 saved
episode/judge artifacts were hash-verified. This is one training seed and
one evaluation panel; no Prompted OPD significance claim or additional
partner, judge, or AgentSense evaluation is made.

\subsection{Model Roles}
\begin{table}[ht]
\centering\small
\caption{\textbf{Model roles in TACT.} Specialists share a frozen backbone; the IG scorer and distillation teacher are distinct roles.}
\label{tab:model-roles}
\begin{tabular}{@{}lp{0.67\linewidth}@{}}
\toprule
Role & Model and information used \\
\midrule
Candidate specialists & Frozen Qwen3.5-27B with expression or strategy instructions. \\
IG scorer & Batch-frozen Qwen3.5-4B student; scores the target goal before and after the action and partner reply. \\
Training/branch partner & Same batch-frozen student snapshot in the partner role. \\
Distillation teacher & Frozen Qwen3.5-27B; conditions on the selected reference and scores original student tokens. \\
Evaluation partner & Fixed initial Qwen3.5-4B. \\
Outcome judge & DeepSeek-v4-pro; evaluates completed dialogues with the seven-dimensional SOTOPIA rubric. \\
\bottomrule
\end{tabular}
\end{table}

\clearpage
\section{AgentSense Transfer Evaluation}
\label{app:agentsense}

\paragraph{Subset and unit of evaluation.}
The released AgentSense data contain 1,225 instances, including 730
two-person instances from 146 templates. We uniformly sample 100 of these
templates without replacement using seed 20260924, before collecting
the reported outcomes, and include all five official instances per
selected template. An instance is a supplied scene record specifying
the characters, their profiles, goals, and private information; the five
instances are not five generated repetitions of one record. The subset
covers 100/146 eligible two-person templates (68.5\%); sampling is by
template, with no additional semantic-category stratification. The fixed
case manifest and selected template IDs accompany the result artifacts.

Each model generates one dialogue per instance. One focal role is fixed
per instance and shared across methods, with 250 instances for each
role. Within each template, the five focal roles split 3/2, and the
majority role is balanced across templates. Starting roles are also
balanced 250/250 and fixed across methods. Initial, concise prompting,
vanilla OPD, and the selected 2,970-node TACT checkpoint each complete
all 500 instances; the frozen initial Qwen3.5-4B always plays the other
role. These checkpoints were fixed before this transfer evaluation.

\paragraph{Interaction protocol.}
Agents receive AgentSense's native role prompts with only their own
profile, goals, and private information. Concise prompting adds the
same baseline instruction to the focal role only. History follows the
native name-prefix and first-line transformation. An empty-history
start instruction is control input and is not counted as a spoken
message. Following SocialRL's released evaluation, the conversation
ends after at most 20 generated messages (ten per role) or immediately
when a generated response contains \texttt{[LEAVE]}. There is no scripted
opening utterance. Non-thinking generation uses temperature 1,
top-$p=1$, no top-$k$ filtering, and a maximum of 128 tokens per
utterance. Repetition penalty is 1 and presence/frequency penalties
are 0. Per-case, per-role, per-turn seeds follow the fixed seed-20260924
schedule. These native decoding settings and the fixed dyadic partner
arrangement define an adapted transfer protocol, rather than a
reproduction of SocialRL's full benchmark evaluation.

\paragraph{Scoring and costs.}
We reuse SocialRL's released holistic goal-success and relationship-change
prompts (source revision \texttt{aa88db035a35}). The judge receives the
saved utterances with the leave marker removed. Goal judges the focal
role's goal list as a whole with one binary score; it is not an average
over separately judged goals. Relationship change estimates the other
character's change in favorability toward the focal character on
$[-1,1]$. We request DeepSeek-v4-flash in non-thinking mode at temperature
0.1 with an output limit of 2,048 tokens; saved responses identify the
API model as \texttt{deepseek-flash}. We retain one valid response per
metric and instance, with bounded transport retries and at most two
validity passes. All 4,000 metric scores are valid; one TACT goal
rating required the second validity pass. Failed API calls never become
zero scores. This evaluation uses these two metrics, without AgentSense's
separate implicit-information question task.

Tokens sums the focal agent's complete generated token IDs, including
formatting and termination tokens, before history transformation.
Messages counts both participants' generated utterances, including
the final leave response. For every method, scores and costs are computed
from the same 500 saved dialogues. Initial, concise prompting, and TACT
were generated on an L20; vanilla OPD was distributed across four A800
replicas with 125 disjoint cases each. Model and tokenizer hashes,
prompts, seed schedules, and decoding settings were verified across
hosts; differing hardware and batching do not imply bitwise-identical
generation. Every dialogue and judge record was joined by case ID and
verified against its source hashes.

\paragraph{Uncertainty and interpretation.}
The five instances of each template form a cluster. We resample the
100 templates with replacement 10,000 times (seed 20260925), retaining
all five instances and using the same resampled clusters for every
method. Tables~\ref{tab:agentsense-ci} and~\ref{tab:agentsense-paired}
report percentile 95\% intervals. These quantify template-sample
uncertainty conditional on the recorded checkpoints, single rollouts,
and judge responses; they do not measure training-seed or judge-repeat
variability. The transfer evidence is restricted to the sampled
two-person subset. The Goal interval for TACT minus vanilla OPD spans
zero; this does not establish equivalence or non-inferiority.

\begin{table}[ht]
\centering\scriptsize\setlength{\tabcolsep}{2pt}
\caption{\textbf{AgentSense means and 95\% template-cluster intervals.}
All four panels contain 500 instances in the same 100 templates.}
\label{tab:agentsense-ci}
\begin{tabular*}{\linewidth}{@{\extracolsep{\fill}}lrrrr@{}}
\toprule
Method & Goal (\%) $\uparrow$ & Rel. $\uparrow$ & Tokens $\downarrow$ & Messages $\downarrow$ \\
\midrule
Initial & 48.4 [42.4, 54.2] & 0.391 [0.330, 0.452] & 899.7 [844.7, 950.9] & 14.25 [13.40, 15.04] \\
Concise prompt & 45.2 [39.4, 51.0] & 0.328 [0.265, 0.388] & 621.6 [574.7, 667.1] & 9.92 [9.18, 10.63] \\
Vanilla OPD & 54.6 [48.8, 60.4] & 0.435 [0.374, 0.492] & 841.8 [791.0, 890.9] & 13.27 [12.48, 14.04] \\
TACT & 54.2 [48.0, 60.2] & 0.432 [0.370, 0.487] & 789.6 [736.4, 842.3] & 12.47 [11.63, 13.30] \\
\bottomrule
\end{tabular*}
\end{table}

\begin{table}[ht]
\centering\scriptsize\setlength{\tabcolsep}{2pt}
\caption{\textbf{Paired AgentSense differences: TACT minus each comparator.}
Goal differences are percentage points; relationship differences retain
the native scale. Brackets give 95\% template-cluster intervals.}
\label{tab:agentsense-paired}
\begin{tabular*}{\linewidth}{@{\extracolsep{\fill}}lrrrr@{}}
\toprule
Comparator & $\Delta$Goal (pp) $\uparrow$ & $\Delta$Rel. $\uparrow$ & $\Delta$Tokens $\downarrow$ & $\Delta$Messages $\downarrow$ \\
\midrule
Initial & 5.8 [0.8, 10.8] & 0.040 [-0.001, 0.081] & -110.1 [-156.9, -62.1] & -1.78 [-2.52, -1.02] \\
Concise prompt & 9.0 [3.8, 14.2] & 0.104 [0.063, 0.145] & 168.0 [130.6, 206.2] & 2.56 [1.97, 3.16] \\
Vanilla OPD & -0.4 [-4.8, 4.2] & -0.003 [-0.040, 0.034] & -52.2 [-91.1, -13.3] & -0.80 [-1.42, -0.18] \\
\bottomrule
\end{tabular*}
\end{table}

\section{Additional Experimental Results}
\label{app:result-details}
\subsection{Full Social Performance and Communication Cost}
\label{app:full-results}
Tables~\ref{tab:main-results-full} and~\ref{tab:ablations-full} expand
Tables~\ref{tab:main-results} and~\ref{tab:ablations} to all seven SOTOPIA
dimensions. They use the same checkpoints, completed 450-case All and
70-case Hard panels, and dialogue-level costs as their main-text counterparts.
Avg is the arithmetic mean of all seven scores on their native scales;
Tokens counts the target agent's generated action tokens, and Turns counts
both participants' environment turns. Metric definitions and ranges appear
in Table~\ref{tab:metric-ranges}. Bold and underlined values mark the best
and second-best distinct displayed values within each panel.

\begin{table}[t]
\centering\scriptsize\setlength{\tabcolsep}{1.0pt}
\caption{\textbf{Full social performance and communication cost on SOTOPIA.} Full-dimensional results for Table~\ref{tab:main-results}. Bold and underlined values indicate the best and second-best results, respectively.}
\label{tab:main-results-full}
\begin{tabular*}{\linewidth}{@{\extracolsep{\fill}}lrrrrrrrrrr@{}}
\toprule
Method & Goal $\uparrow$ & Rel. $\uparrow$ & Kno. $\uparrow$ & Bel. $\uparrow$ & Sec. $\uparrow$ & Rules $\uparrow$ & Fin. $\uparrow$ & Avg $\uparrow$ & Tokens $\downarrow$ & Turns $\downarrow$ \\
\midrule
\multicolumn{11}{@{}l}{\textit{SOTOPIA-All ($n=450$)}}\\
Initial student & 4.327 & -0.193 & 3.707 & 7.800 & -0.442 & -0.669 & 0.102 & 2.090 & 300.3 & 13.06 \\
Concise prompt & 4.691 & -0.009 & 3.687 & 7.827 & -0.311 & -0.476 & 0.216 & 2.232 & \textbf{235.0} & 13.04 \\
Teacher SFT & 4.840 & 0.476 & 3.944 & \underline{8.431} & -0.213 & -0.251 & 0.367 & 2.513 & 293.4 & \underline{10.95} \\
Vanilla OPD & 4.960 & 0.580 & 4.060 & 8.376 & \underline{-0.171} & -0.224 & 0.307 & 2.555 & 275.4 & 11.10 \\
Prompted OPD & 5.151 & 0.680 & 4.122 & 8.413 & \textbf{-0.160} & \underline{-0.207} & \underline{0.420} & 2.631 & \underline{275.1} & \textbf{10.61} \\
SFT+SDPO & \underline{5.276} & \textbf{1.489} & \textbf{4.300} & \textbf{8.460} & -0.187 & \textbf{-0.120} & 0.333 & \textbf{2.793} & 435.7 & 14.40 \\
Sotopia-RL & 4.791 & \underline{0.976} & \underline{4.253} & 7.253 & -0.884 & -0.422 & 0.104 & 2.296 & 1104.1 & 16.88 \\
TACT & \textbf{5.611} & 0.818 & 4.171 & 8.356 & -0.278 & -0.229 & \textbf{0.527} & \underline{2.711} & 280.6 & 14.22 \\
\midrule
\multicolumn{11}{@{}l}{\textit{SOTOPIA-Hard ($n=70$)}}\\
Initial student & 3.457 & -0.829 & 3.414 & 7.729 & -0.471 & -1.057 & -0.200 & 1.720 & 254.9 & 11.26 \\
Concise prompt & \underline{3.857} & -0.986 & 3.386 & 7.471 & -0.257 & -1.243 & 0.229 & 1.780 & \textbf{186.0} & 10.23 \\
Teacher SFT & 3.757 & -0.600 & 3.671 & \textbf{8.371} & -0.386 & -0.700 & 0.329 & 2.063 & 238.2 & 8.39 \\
Vanilla OPD & 3.686 & -0.586 & 3.614 & \underline{8.329} & -0.300 & -0.629 & 0.243 & 2.051 & \underline{201.1} & \textbf{8.03} \\
Prompted OPD & 3.686 & -0.586 & 3.729 & 8.229 & \textbf{-0.114} & \underline{-0.571} & \underline{0.357} & 2.104 & 220.0 & \underline{8.07} \\
SFT+SDPO & 3.543 & \textbf{0.343} & \textbf{4.086} & 8.200 & \textbf{-0.114} & \textbf{-0.229} & -0.129 & \textbf{2.243} & 315.4 & 10.99 \\
Sotopia-RL & 3.000 & \underline{0.143} & \underline{3.929} & 6.943 & -0.557 & -0.957 & -0.543 & 1.708 & 1029.7 & 15.63 \\
TACT & \textbf{4.371} & -0.486 & 3.657 & 8.143 & \underline{-0.200} & \underline{-0.571} & \textbf{0.471} & \underline{2.198} & 233.7 & 11.67 \\
\bottomrule
\end{tabular*}
\end{table}

\begin{table}[t]
\centering\scriptsize\setlength{\tabcolsep}{1.4pt}
\caption{\textbf{Full specialist, selection, and supervision comparisons.} Full-dimensional results for Table~\ref{tab:ablations}. TACT uses the same selected 2,970-node checkpoint and original evaluation cohort as Table~\ref{tab:main-results}; checkpoint details appear in Appendix~\ref{app:selected-tact}. All populated rows use complete 450/70 panels after missing-case recovery. No reference retains candidate selection but removes selected-reference conditioning from the scoring teacher; its checkpoint uses 2,818 training nodes, 74 updates, and learning rate $10^{-5}$. Intervals for the full-method and single-specialist rows appear in Appendix~\ref{app:complete-statistics}.}
\label{tab:ablations-full}
\begin{tabular*}{\linewidth}{@{\extracolsep{\fill}}lrrrrrrrrrr@{}}
\toprule
Method & Goal $\uparrow$ & Rel. $\uparrow$ & Kno. $\uparrow$ & Bel. $\uparrow$ & Sec. $\uparrow$ & Rules $\uparrow$ & Fin. $\uparrow$ & Avg $\uparrow$ & Tokens $\downarrow$ & Turns $\downarrow$ \\
\midrule
\multicolumn{11}{@{}l}{\textit{SOTOPIA-All ($n=450$)}}\\
Expression only & \underline{5.327} & \textbf{0.869} & \underline{4.082} & 8.289 & \underline{-0.147} & -0.338 & \underline{0.433} & \underline{2.645} & \textbf{250.1} & 14.60 \\
Strategy only & 5.073 & 0.764 & 4.067 & 8.218 & -0.162 & -0.238 & 0.356 & 2.583 & 310.6 & 13.02 \\
IG-only ranking & 5.093 & 0.631 & 3.987 & \textbf{8.384} & \underline{-0.147} & \textbf{-0.213} & 0.373 & 2.587 & 279.8 & \underline{12.57} \\
Token-only ranking & 5.009 & 0.611 & 3.978 & 8.216 & -0.162 & -0.289 & 0.331 & 2.528 & 256.2 & 13.17 \\
Random ranking & 5.049 & 0.613 & 4.027 & 8.276 & \textbf{-0.136} & -0.258 & 0.351 & 2.560 & \underline{252.0} & 12.74 \\
No reference & 4.860 & 0.620 & \underline{4.082} & 8.338 & -0.151 & -0.238 & 0.291 & 2.543 & 297.6 & \textbf{11.78} \\
TACT & \textbf{5.611} & \underline{0.818} & \textbf{4.171} & \underline{8.356} & -0.278 & \underline{-0.229} & \textbf{0.527} & \textbf{2.711} & 280.6 & 14.22 \\
\midrule
\multicolumn{11}{@{}l}{\textit{SOTOPIA-Hard ($n=70$)}}\\
Expression only & 3.800 & \underline{-0.386} & 3.429 & 7.857 & -0.171 & -0.814 & 0.057 & 1.967 & 212.1 & 12.41 \\
Strategy only & 3.957 & \textbf{-0.257} & 3.543 & 7.900 & \underline{-0.143} & -0.700 & 0.386 & 2.098 & 236.1 & 9.79 \\
IG-only ranking & \underline{4.043} & -0.643 & 3.500 & \underline{8.200} & -0.157 & \textbf{-0.486} & \underline{0.414} & \underline{2.124} & 220.4 & 9.66 \\
Token-only ranking & 3.914 & -0.714 & 3.429 & 7.886 & -0.186 & -0.829 & 0.257 & 1.965 & \underline{202.4} & 10.39 \\
Random ranking & 3.600 & -0.771 & 3.400 & 8.000 & \textbf{-0.129} & -0.629 & 0.286 & 1.965 & \textbf{190.8} & \underline{9.34} \\
No reference & 3.386 & -0.686 & \underline{3.614} & \textbf{8.271} & -0.257 & \underline{-0.571} & 0.057 & 1.973 & 227.0 & \textbf{8.93} \\
TACT & \textbf{4.371} & -0.486 & \textbf{3.657} & 8.143 & -0.200 & \underline{-0.571} & \textbf{0.471} & \textbf{2.198} & 233.7 & 11.67 \\
\bottomrule
\end{tabular*}
\end{table}

IG-only, Token-only, and Random each train a fresh Qwen3.5-4B student over
the same 1,129 training configurations once, with learning rate $10^{-5}$,
seed 20360915, and 71 updates. Their final checkpoints contain 3,323,
3,843, and 3,174 admitted training nodes, respectively. Each uses the
original eligibility gates and selected-reference conditioning;
Token-only minimizes the full action-token count among eligible candidates,
so positive-IG admission remains in effect. Only their final checkpoints
are evaluated, whereas TACT uses the selected 2,970-node checkpoint.
The two new panels use the same original 450-case manifest, frozen initial
4B partner, and DeepSeek-v4-pro scoring protocol. Each has 446 first-pass
scores: IG-only adds two same-request retries and two schema clarifications;
Token-only adds four same-request retries. All use the original saved
dialogues and finish with 450 valid scores, including all 70 Hard cases.

\subsection{Complete-Panel Uncertainty and Paired Differences}
\label{app:complete-statistics}
We resample benchmark scenarios with replacement, retaining all five character pairings within each sampled scenario. All uses 90 scenario clusters (450 cases); Hard uses its 14 clusters (70 cases). We use 10,000 resamples with seed 20260924 and report percentile 95\% intervals. Paired comparisons use identical resampled scenarios and matched cases for both methods. These pointwise intervals quantify scenario-sampling uncertainty conditional on the recorded runs, recovery procedures, and checkpoints; they are not standard deviations across training seeds, repeated-judge uncertainty, or selection-adjusted intervals. No multiple-comparison adjustment is applied.

\begin{table}[t]
\centering\scriptsize\setlength{\tabcolsep}{2pt}
\caption{\textbf{95\% intervals for complete-panel means.} Full point estimates appear in Tables~\ref{tab:main-results-full} and~\ref{tab:ablations-full}.}
\label{tab:complete-intervals}
\begin{tabular*}{\linewidth}{@{\extracolsep{\fill}}lrrrrrr@{}}
\toprule
Method & Goal $\uparrow$ & Rel. $\uparrow$ & Kno. $\uparrow$ & Avg $\uparrow$ & Tokens $\downarrow$ & Turns $\downarrow$ \\
\midrule
\multicolumn{7}{@{}l}{\textit{SOTOPIA-All}}\\
Initial student & [3.90, 4.77] & [-0.48, 0.09] & [3.45, 3.96] & [1.94, 2.24] & [282.7, 318.7] & [12.30, 13.82] \\
Concise prompt & [4.20, 5.19] & [-0.32, 0.30] & [3.43, 3.95] & [2.07, 2.39] & [222.4, 247.9] & [12.33, 13.75] \\
Teacher SFT & [4.36, 5.32] & [0.19, 0.75] & [3.66, 4.22] & [2.38, 2.65] & [276.2, 310.5] & [10.26, 11.66] \\
Vanilla OPD & [4.46, 5.46] & [0.27, 0.89] & [3.78, 4.34] & [2.41, 2.70] & [257.0, 294.3] & [10.29, 11.93] \\
SFT+SDPO & [4.77, 5.79] & [1.21, 1.77] & [4.05, 4.54] & [2.66, 2.93] & [381.4, 523.0] & [13.65, 15.14] \\
Sotopia-RL & [4.36, 5.23] & [0.69, 1.26] & [4.04, 4.47] & [2.13, 2.46] & [1056.1, 1152.0] & [16.31, 17.42] \\
TACT & [5.13, 6.09] & [0.51, 1.13] & [3.90, 4.44] & [2.56, 2.85] & [264.2, 296.9] & [13.39, 15.03] \\
Expression only & [4.82, 5.83] & [0.56, 1.17] & [3.80, 4.36] & [2.50, 2.79] & [237.8, 262.6] & [13.88, 15.31] \\
Strategy only & [4.58, 5.57] & [0.46, 1.06] & [3.80, 4.33] & [2.44, 2.72] & [290.6, 331.2] & [12.16, 13.88] \\
\midrule
\multicolumn{7}{@{}l}{\textit{SOTOPIA-Hard}}\\
Initial student & [2.26, 4.83] & [-1.54, -0.10] & [2.77, 4.07] & [1.25, 2.17] & [213.6, 299.4] & [9.14, 13.54] \\
Concise prompt & [2.46, 5.44] & [-1.79, -0.07] & [2.81, 3.97] & [1.29, 2.26] & [151.7, 226.6] & [8.26, 12.50] \\
Teacher SFT & [2.70, 4.96] & [-1.20, -0.06] & [2.94, 4.39] & [1.64, 2.39] & [198.9, 276.6] & [7.07, 9.80] \\
Vanilla OPD & [2.33, 5.21] & [-1.21, -0.03] & [2.90, 4.34] & [1.63, 2.46] & [171.7, 229.4] & [6.71, 9.43] \\
SFT+SDPO & [2.33, 4.91] & [-0.21, 0.89] & [3.56, 4.70] & [1.98, 2.54] & [258.6, 369.3] & [9.17, 12.86] \\
Sotopia-RL & [2.14, 4.07] & [-0.60, 0.81] & [3.63, 4.24] & [1.31, 2.09] & [868.3, 1189.2] & [13.91, 17.29] \\
TACT & [3.00, 5.80] & [-1.26, 0.31] & [3.01, 4.27] & [1.73, 2.62] & [198.2, 272.9] & [9.59, 14.03] \\
Expression only & [2.47, 5.26] & [-1.13, 0.27] & [2.71, 4.13] & [1.52, 2.33] & [178.9, 248.6] & [10.26, 14.67] \\
Strategy only & [2.66, 5.46] & [-0.89, 0.29] & [2.90, 4.14] & [1.69, 2.45] & [199.0, 280.2] & [8.04, 11.89] \\
\bottomrule
\end{tabular*}
\end{table}

\begin{table}[t]
\centering\scriptsize\setlength{\tabcolsep}{2pt}
\caption{\textbf{Paired differences: main-table TACT minus each comparator.} Entries are mean differences [95\% scenario-cluster interval]. Positive Goal and negative cost differences favor TACT. Comparators retain their recorded training pipelines.}
\label{tab:complete-paired}
\begin{tabular*}{\linewidth}{@{\extracolsep{\fill}}lrrr@{}}
\toprule
Comparator & $\Delta$ Goal $\uparrow$ & $\Delta$ Tokens $\downarrow$ & $\Delta$ Turns $\downarrow$ \\
\midrule
\multicolumn{4}{@{}l}{\textit{SOTOPIA-All}}\\
Initial student & +1.284 [0.947, 1.631] & -19.7 [-33.8, -5.5] & +1.158 [0.573, 1.738] \\
Concise prompt & +0.920 [0.616, 1.231] & +45.6 [33.6, 57.6] & +1.182 [0.609, 1.742] \\
Teacher SFT & +0.771 [0.456, 1.107] & -12.8 [-26.7, 1.0] & +3.269 [2.669, 3.873] \\
Vanilla OPD & +0.651 [0.331, 0.976] & +5.2 [-8.0, 18.9] & +3.122 [2.540, 3.720] \\
SFT+SDPO & +0.336 [0.027, 0.647] & -155.1 [-234.9, -107.2] & -0.182 [-0.805, 0.413] \\
Sotopia-RL & +0.820 [0.469, 1.167] & -823.5 [-868.8, -779.4] & -2.653 [-3.362, -1.976] \\
Expression only & +0.284 [0.002, 0.562] & +30.5 [19.8, 41.2] & -0.378 [-0.933, 0.169] \\
Strategy only & +0.538 [0.251, 0.831] & -29.9 [-44.0, -15.6] & +1.207 [0.644, 1.773] \\
\midrule
\multicolumn{4}{@{}l}{\textit{SOTOPIA-Hard}}\\
Initial student & +0.914 [-0.043, 1.943] & -21.2 [-42.9, 0.8] & +0.414 [-0.886, 1.700] \\
Concise prompt & +0.514 [-0.457, 1.500] & +47.6 [22.2, 72.9] & +1.443 [-0.043, 3.071] \\
Teacher SFT & +0.614 [-0.186, 1.514] & -4.5 [-30.3, 23.6] & +3.286 [1.914, 4.814] \\
Vanilla OPD & +0.686 [-0.114, 1.657] & +32.5 [4.2, 61.8] & +3.643 [2.057, 5.371] \\
SFT+SDPO & +0.829 [0.157, 1.700] & -81.8 [-129.5, -38.5] & +0.686 [-0.857, 2.343] \\
Sotopia-RL & +1.371 [0.586, 2.186] & -796.1 [-953.3, -643.5] & -3.957 [-5.957, -1.857] \\
Expression only & +0.571 [-0.043, 1.229] & +21.6 [5.6, 37.3] & -0.743 [-1.943, 0.514] \\
Strategy only & +0.414 [-0.229, 1.100] & -2.5 [-23.2, 19.6] & +1.886 [0.929, 2.986] \\
\bottomrule
\end{tabular*}
\end{table}

\subsection{Development Endpoints and Main-Table Reference}
\begin{table}[t]
\centering\footnotesize\setlength{\tabcolsep}{2pt}
\caption{\textbf{Development endpoints and the selected main-table checkpoint.} Development rows use baseline-paired valid cases from the 90-case panel. The $10^{-5}$ rows reproduce the complete 450/70-case evaluation in Table~\ref{tab:main-results-full}. Nodes count admitted training actions at the evaluated checkpoint, not the final size of its training run. Panels have different evaluation coverage and checkpoint selection; no cross-panel ranking is applied.}
\label{tab:development}
\begin{tabular*}{\linewidth}{@{\extracolsep{\fill}}lrrrrrrrrrr@{}}
\toprule
& \multicolumn{8}{c}{Social performance $\uparrow$} & \multicolumn{2}{c}{Communication cost $\downarrow$} \\
\cmidrule(lr){2-9}\cmidrule(l){10-11}
LR / nodes & Goal & Rel. & Kno. & Bel. & Sec. & Rules & Fin. & Avg & Tokens & Turns \\
\midrule
\multicolumn{11}{@{}l}{\textit{SOTOPIA-All: development ($n=83/84$, respectively)}}\\
$10^{-6}$ / 3,118 & 5.07 & 0.33 & 3.66 & 8.14 & -0.30 & -0.36 & 0.37 & 2.42 & 251.6 & 12.22 \\
$5\!\times\!10^{-6}$ / 3,606 & 5.32 & 0.51 & 3.90 & 8.29 & -0.20 & -0.25 & 0.38 & 2.56 & 293.5 & 13.67 \\
\addlinespace
\multicolumn{11}{@{}l}{\textit{SOTOPIA-All: main-table reference ($n=450$)}}\\
$10^{-5}$ / 2,970 & 5.611 & 0.818 & 4.171 & 8.356 & -0.278 & -0.229 & 0.527 & 2.711 & 280.6 & 14.22 \\
\midrule
\multicolumn{11}{@{}l}{\textit{SOTOPIA-Hard: development ($n=13$ for each row)}}\\
$10^{-6}$ / 3,118 & 3.46 & -0.62 & 3.38 & 7.62 & -0.38 & -0.38 & 0.38 & 1.92 & 195.6 & 9.69 \\
$5\!\times\!10^{-6}$ / 3,606 & 3.77 & -1.00 & 3.46 & 7.77 & -0.38 & -0.85 & 0.54 & 1.90 & 190.8 & 9.54 \\
\addlinespace
\multicolumn{11}{@{}l}{\textit{SOTOPIA-Hard: main-table reference ($n=70$)}}\\
$10^{-5}$ / 2,970 & 4.371 & -0.486 & 3.657 & 8.143 & -0.200 & -0.571 & 0.471 & 2.198 & 233.7 & 11.67 \\
\bottomrule
\end{tabular*}
\end{table}

Table~\ref{tab:development} retains the final development endpoints at $10^{-6}$ and $5\times10^{-6}$ and separately identifies the selected $10^{-5}$ main-table checkpoint. The latter's scores and costs come from the same original evaluation cohort as Table~\ref{tab:main-results}.

For the two development rows, All paired baseline Goal scores are 3.928 and 3.881, and paired Goal changes are 1.145 [0.434, 1.855] and 1.440 [0.726, 2.155], respectively (95\% pointwise intervals).
The broader historical development study contains 31 completed checkpoints at $10^{-6}$, 36 at $5\times10^{-6}$, and 29 at $10^{-5}$. Across those 96 evaluations, 8,203 of 8,640 planned case evaluations have valid scores and 8,007 can be paired with the baseline. Each newer training run visits the same ordered-configuration pool once, but yields a different number of admitted nodes; the historical run has a different exposure schedule. Development results neither replace the 450-case results nor establish an untouched test set.

Historical within-run comparisons do not show uniformly improving performance. On their respective common-valid All cohorts, the late-minus-early Goal changes are $-0.405$ (95\% interval $[-0.740,-0.067]$; same context, $n=385$) and $-0.240$ ($[-0.589,0.112]$; reference context, $n=391$). These describe historical trajectories, not the selected main-table checkpoint or independent training seeds.

\subsection{Training Exposure}
\begin{table}[ht]
\centering\small
\caption{\textbf{Training exposure.} Reference-conditioned runs at the reported training endpoints. A selected training node, a dialogue configuration, and an optimizer update are distinct units.}
\label{tab:training-exposure}
\begin{tabular*}{\linewidth}{@{\extracolsep{\fill}}lrrr@{}}
\toprule
Run & Unique configs. & Nodes & Updates \\
\midrule
Reference, $10^{-5}$ & 931 & 3,000 & 59 \\
Reference, $5\times10^{-6}$ & 1,129 & 3,606 & 71 \\
Reference, $10^{-6}$ & 1,129 & 3,118 & 71 \\
\bottomrule
\end{tabular*}
\end{table}
The two newer runs traverse the same configuration pool once. Historical reference training instead includes 944 dialogues from 931 unique configurations. Node counts are outputs of collection and admission, not equal-compute budgets. Training-cost accounting must separate student collection, teacher proposals, partner branches, probability scoring, and updates; comparable aggregate costs have not yet been established.

\clearpage
\subsection{Candidate Selection and Downstream Validation}
\label{app:selection-diagnostics}
\paragraph{Teacher adoption and filtering funnel.}
Tables~\ref{tab:candidate-coverage} and~\ref{tab:selection-opportunities}
summarize the training records of the same $10^{-5}$, 2,970-node
checkpoint used in the main results. They cover all 58 committed OPD
updates (batches 0--57), 928 dialogue-configuration visits, and 6,725
target actions. Expression supplies 2,036 of the 2,970 references used
for OPD (68.6\%), and strategy supplies 934 (31.4\%).
Table~\ref{tab:candidate-coverage} separates this share of training
references from each specialist's proposal-to-selection rate.

\begin{table}[!ht]
\centering\small
\caption{\textbf{Specialist proposal funnel and OPD adoption.} Same 2,970-node checkpoint as the main results. Scorable requires both the original and candidate branches to be scorable. Funnel percentages use each specialist's 6,624 proposals; the final row instead uses all 2,970 references actually used for updates. Counts describe supervision frequency.}
\label{tab:candidate-coverage}
\begin{tabular*}{\linewidth}{@{\extracolsep{\fill}}lrr@{}}
\toprule
Stage & Expression & Strategy \\
\midrule
Proposed & 6,624 (100.0\%) & 6,624 (100.0\%) \\
Scorable & 6,039 (91.2\%) & 5,665 (85.5\%) \\
Eligible & 2,405 (36.3\%) & 1,370 (20.7\%) \\
Selected & 2,036 (30.7\%) & 934 (14.1\%) \\
Used for OPD & 2,036 (30.7\%) & 934 (14.1\%) \\
\midrule
Share of used references & 68.6\% & 31.4\% \\
\bottomrule
\end{tabular*}
\end{table}
The eligibility rules are those in Section~\ref{sec:validation}: both
specialists require positive own-action IG; expression additionally requires
fewer tokens than the original action, and strategy requires larger IG than
the original. All selected references enter updates. Invalid source actions
account for the 101 nodes without proposals. Of the 3,755 nodes with no eligible
candidate, 330 lack a scorable original action/branch and 3,425 have a scorable
original but no candidate passing admission; the 101 invalid-source nodes are
already included in the 330.

\begin{table}[!ht]
\centering\small
\caption{\textbf{Selection opportunities and ranking disagreements.} Eligibility percentages use all 6,725 target actions. Ranking disagreements use only the 805 dual-eligible nodes and count a disagreement when the selected candidate falls outside the alternative rule's best set; tied optima are not disagreements.}
\label{tab:selection-opportunities}
\begin{tabular*}{\linewidth}{@{\extracolsep{\fill}}lrr@{}}
\toprule
Diagnostic & Count & Percentage \\
\midrule
No eligible candidate & 3,755 / 6,725 & 55.8\% \\
Exactly one eligible candidate & 2,165 / 6,725 & 32.2\% \\
Both candidates eligible & 805 / 6,725 & 12.0\% \\
\midrule
Disagreement with IG-only ranking & 165 / 805 & 20.5\% \\
Disagreement with shortest-eligible ranking & 315 / 805 & 39.1\% \\
\bottomrule
\end{tabular*}
\end{table}
Both specialists contribute training references, with expression selected
more often. The ratio reflects both their different admission conditions and
subsequent ranking. Ranking can affect the choice only at the 805 dual-eligible
nodes (27.1\% of the 2,970 selected nodes); elsewhere admission already fixes
the outcome. These logged-candidate diagnostics establish how supervision is
allocated. Assessing its effect on final social outcomes requires the trained
selector comparisons and continuation diagnostics below.

\clearpage
\paragraph{Random selector audit.}
\label{app:random-selection-audit}
The Random run in Table~\ref{tab:ablations} traverses 1,129 training
configurations once and records 7,163 target-action nodes. Of these, 3,989
have no eligible candidate (including 498 nodes without a valid selection
stage), 2,333 have exactly one, and 841 have two. All 3,174 selected
references enter the 71 committed OPD updates; expression supplies 2,192
(69.06\%) and strategy 982 (30.94\%). Counts are deduplicated against the
committed training ledger, including the recovered batch.

At the 841 dual-eligible nodes, Random chooses expression 438 times and
strategy 403 times. It selects the strictly lower-IG/token candidate at
419 nodes: 49.82\% of dual-eligible nodes, 13.20\% of all 3,174 training
nodes, and 5.85\% of all 7,163 collected nodes. There are no exact
IG/token ties. The other 2,333 training nodes have only one eligible
candidate, so the ranking rule cannot change their reference. Of the
419 lower-efficiency selections, 168 (5.29\% of training nodes) are
Pareto-dominated: the alternative has at least as much IG and no more
action tokens, with at least one strict inequality. These are local
disagreements with the original selector on Random's own trajectory,
not demonstrated downstream errors or a counterfactual replay of TACT
training. The observed final-model difference also includes different
training exposure and checkpoint selection.

\paragraph{Local IG gains and downstream goal attainment.}
\label{app:continuation-diagnostics}
TACT uses local goal-support feedback to evaluate teacher revisions. We examine
how candidate--student IG gains relate to final outcomes by replaying historical
training prefixes and independently continuing the original and revised actions.
The primary cohort contains 300 randomly sampled committed OPD nodes, with
100 from each third of the first 2,970 training nodes. The same historical
pre-update snapshot is used for both interacting roles at each node; the
prefix, scene, profiles, original action, teacher candidates, and recorded IG
are held fixed. Each distinct legal action receives two fresh continuations
at temperature 1 with the native 20-turn, two-stale-turn, or leave termination
rules. These are training-node diagnostics rather than the SOTOPIA-All/Hard
benchmark panels. Final dialogues are rated using the main-table DeepSeek judge
and the same fixed target role.

For a teacher candidate $c$ and the original student action $s$, we compute
\[
\Delta\mathrm{IG}=\mathrm{IG}(c)-\mathrm{IG}(s),\qquad
\Delta\mathrm{Goal}=\overline{\mathrm{Goal}}(c)-\overline{\mathrm{Goal}}(s),
\]
where each bar averages two independent continuations. A pair is included
when both actions have two valid ratings and recorded IG. The resulting
351 pairs come from 218 nodes in 125 scenarios. In this post-hoc grouping,
positive-gain pairs have $\Delta\mathrm{IG}>0$. High-gain pairs are in the
upper quartile of positive gains within each training phase. Thresholds use
all available historical IG pairs in the random cohort, including those with
missing continuation outcomes, and are 0.015663, 0.018207, and 0.017939 for
early, middle, and late training. This is a diagnostic grouping, distinct from
TACT's own-action-IG admission rules. We average paired differences with equal
weight per pair and use 10,000 scenario-cluster bootstrap resamples for the
reported intervals.

\begin{table}[!ht]
\centering\small
\caption{\textbf{Candidate--student IG gains and complete-continuation outcomes.}
Exploratory groups within the random 300-node cohort. Goal changes compare a
teacher candidate with the original student action; brackets give 95\%
scenario-cluster intervals. Counts are paired comparisons, not independent
training runs; one node may contribute two pairs.}
\label{tab:downstream-validation}
\begin{tabular*}{\linewidth}{@{\extracolsep{\fill}}lrrr@{}}
\toprule
Candidate group & Pairs & Nodes & $\Delta$Goal [95\% interval] \\
\midrule
All comparable candidates & 351 & 218 & $+0.103$ [$-0.012$, $+0.223$] \\
Positive IG gain & 187 & 149 & $+0.147$ [$+0.017$, $+0.277$] \\
Upper quartile of positive gains & 45 & 38 & $+0.244$ [$-0.073$, $+0.605$] \\
Remaining positive gains & 142 & 122 & $+0.116$ [$-0.021$, $+0.255$] \\
\bottomrule
\end{tabular*}
\end{table}

\paragraph{Positive local signal.}
Positive-IG-gain candidates yield an observed mean Goal improvement of
$+0.147$ over the original student action (95\% interval $[+0.017,+0.277]$).
The upper-quartile group has a larger observed mean improvement of $+0.244$,
compared with $+0.116$ for the remaining positive-gain candidates
(Table~\ref{tab:downstream-validation}). These observations provide a positive
local signal: candidates with higher IG than the student's original action
achieve better average downstream Goal in the analyzed positive-gain pairs.
Across all comparable pairs, the rank association is $\rho=0.030$
($[-0.088,+0.144]$). We therefore distinguish average gains within the
positive-gain group from a monotonic calibration of IG to final Goal.

\paragraph{Supplementary selection and gap analyses.}
A separate 100-node cohort samples large \emph{teacher--teacher} IG gaps,
$|\mathrm{IG}(c_{\mathrm{expr}})-\mathrm{IG}(c_{\mathrm{strat}})|$, using each
phase's upper quartile. This definition differs from the candidate--student
gain grouping above. The two cohorts are disjoint and exclude an earlier
200-node pilot, whose one-continuation results are not pooled here.
Together they contain 1,174 distinct legal actions and 2,348 continuation
attempts: 2,036 have valid final scores and 312 terminate with invalid action
formats. Of the valid scores, 1,864 are obtained in the first pass and 172
through bounded clarification of schema-only judge outputs; the original
valid scores are preserved. Missing continuations are not assigned zero.

Table~\ref{tab:continuation-selectors} evaluates the original saved selectors
on common complete nodes: all legal branches must have both ratings, leaving
153 random-cohort nodes and 43 teacher-gap-cohort nodes. Random is the uniform
expectation over eligible candidates. The shortest-legal rule also includes
the original action; IG-only and TACT retain the historical eligible set.
These comparisons reuse branch outcomes and do not retrain the selectors.
Tokens and turns count the target's generated action tokens and remaining
environment turns from the intervention onward, including the intervention.
Each node has equal weight within its cohort.

\begin{table}[!ht]
\centering\small\setlength{\tabcolsep}{3pt}
\caption{\textbf{Selector outcomes on common complete training nodes.}
Differences are relative to freshly continuing the original student action.
Goal brackets are 95\% scenario-cluster intervals; cost columns are mean
differences. Positive Goal and negative cost differences favor the selector.}
\label{tab:continuation-selectors}
\begin{tabular*}{\linewidth}{@{\extracolsep{\fill}}lrrr@{}}
\toprule
Selection & $\Delta$Goal [95\% interval] & $\Delta$Tokens & $\Delta$Turns \\
\midrule
\multicolumn{4}{@{}l}{\textit{Randomly sampled nodes ($n=153$)}}\\
Shortest legal action & $-0.023$ [$-0.159$, $+0.114$] & $-17.03$ & $-0.157$ \\
IG-only, eligible & $+0.023$ [$-0.101$, $+0.145$] & $-9.20$ & $+0.020$ \\
Random, eligible & $+0.025$ [$-0.103$, $+0.149$] & $-10.35$ & $-0.026$ \\
TACT selection & $+0.020$ [$-0.106$, $+0.144$] & $-9.19$ & $+0.065$ \\
\midrule
\multicolumn{4}{@{}l}{\textit{Large teacher--teacher IG gaps ($n=43$)}}\\
Shortest legal action & $-0.291$ [$-0.786$, $+0.071$] & $-22.35$ & $-0.558$ \\
IG-only, eligible & $-0.244$ [$-0.744$, $+0.128$] & $-4.37$ & $+0.023$ \\
Random, eligible & $-0.291$ [$-0.786$, $+0.074$] & $-10.29$ & $-0.192$ \\
TACT selection & $-0.256$ [$-0.756$, $+0.116$] & $-5.24$ & $-0.047$ \\
\bottomrule
\end{tabular*}
\end{table}

On random-cohort common nodes, TACT and Random have closely matched observed
Goal changes: their paired difference is $-0.005$ ($[-0.025,+0.016]$).
Table~\ref{tab:continuation-correlations} reports the corresponding
candidate--student and teacher--teacher associations on each comparison's
available support. In the large teacher-gap cohort, eligible candidate--student
pairs exhibit a positive rank association ($\rho=0.256$). The cohort definitions
and comparison units are kept separate throughout.

\begin{table}[!ht]
\centering\small\setlength{\tabcolsep}{3pt}
\caption{\textbf{IG--Goal rank associations in the two continuation cohorts.}
Spearman correlations use paired differences, with fixed expression-minus-strategy
direction for teacher--teacher comparisons. Brackets give pointwise 95\%
scenario-cluster intervals; exploratory comparisons are not multiplicity-adjusted.
Each row uses the complete pairs required for that comparison.}
\label{tab:continuation-correlations}
\begin{tabular*}{\linewidth}{@{\extracolsep{\fill}}lrrr@{}}
\toprule
Comparison & Pairs & Nodes & Spearman $\rho$ [95\% interval] \\
\midrule
\multicolumn{4}{@{}l}{\textit{Randomly sampled nodes}}\\
Candidate minus student & 351 & 218 & $+0.030$ [$-0.088$, $+0.144$] \\
Eligible candidate minus student & 230 & 193 & $+0.084$ [$-0.048$, $+0.212$] \\
Expression minus strategy & 163 & 163 & $-0.001$ [$-0.163$, $+0.163$] \\
\midrule
\multicolumn{4}{@{}l}{\textit{Large teacher--teacher IG gaps}}\\
Candidate minus student & 111 & 68 & $+0.002$ [$-0.195$, $+0.202$] \\
Eligible candidate minus student & 67 & 57 & $+0.256$ [$+0.030$, $+0.484$] \\
Expression minus strategy & 49 & 49 & $-0.160$ [$-0.461$, $+0.143$] \\
\bottomrule
\end{tabular*}
\end{table}

\subsection{Partner Generalization and Judge Sensitivity}
\label{app:robustness-details}
\begin{table}[!t]
\centering\footnotesize\setlength{\tabcolsep}{1.4pt}
\caption{\textbf{Sensitivity to the interaction partner.} The same four methods are compared with the initial Qwen3.5-4B partner, Llama-3.1-8B-Instruct, and a partner using the evaluated policy (self-play). Each block uses the same 450 scenario--character configurations, including 70 Hard cases, and DeepSeek-v4-pro scoring. Fixed-initial results are reproduced from Tables~\ref{tab:main-results} and~\ref{tab:main-results-full}. Bold and underlining mark the best and second-best means among the four methods within each partner--subset block.}
\label{tab:partner-comparison}
\label{tab:robustness}
\label{tab:selfplay}
\begin{tabular*}{\linewidth}{@{\extracolsep{\fill}}lrrrrrrrrrr@{}}
\toprule
Method & Goal $\uparrow$ & Rel. $\uparrow$ & Kno. $\uparrow$ & Bel. $\uparrow$ & Sec. $\uparrow$ & Rules $\uparrow$ & Fin. $\uparrow$ & Avg $\uparrow$ & Tokens $\downarrow$ & Turns $\downarrow$ \\
\midrule
\multicolumn{11}{@{}l}{\textit{SOTOPIA-All ($n=450$)} --- Fixed initial Qwen3.5-4B}\\
Initial & 4.327 & -0.193 & 3.707 & 7.800 & -0.442 & -0.669 & 0.102 & 2.090 & 300.3 & 13.06 \\
Concise prompt & 4.691 & -0.009 & 3.687 & 7.827 & -0.311 & -0.476 & 0.216 & 2.232 & \textbf{235.0} & \underline{13.04} \\
Vanilla OPD & \underline{4.960} & \underline{0.580} & \underline{4.060} & \textbf{8.376} & \textbf{-0.171} & \textbf{-0.224} & \underline{0.307} & \underline{2.555} & \underline{275.4} & \textbf{11.10} \\
TACT & \textbf{5.611} & \textbf{0.818} & \textbf{4.171} & \underline{8.356} & \underline{-0.278} & \underline{-0.229} & \textbf{0.527} & \textbf{2.711} & 280.6 & 14.22 \\
\midrule
\multicolumn{11}{@{}l}{\textit{SOTOPIA-All ($n=450$)} --- Fixed Llama-3.1-8B-Instruct}\\
Initial & 4.987 & 0.153 & 3.727 & 7.956 & -0.393 & -0.576 & 0.198 & 2.293 & 357.0 & 14.99 \\
Concise prompt & 4.682 & -0.058 & 3.620 & 7.736 & -0.242 & -0.658 & 0.171 & 2.179 & \textbf{267.4} & \underline{14.52} \\
Vanilla OPD & \underline{5.547} & \underline{0.884} & \underline{4.193} & \textbf{8.358} & \underline{-0.189} & \textbf{-0.251} & \underline{0.498} & \underline{2.720} & 361.4 & \textbf{13.74} \\
TACT & \textbf{5.807} & \textbf{1.056} & \textbf{4.209} & \underline{8.300} & \textbf{-0.147} & \underline{-0.264} & \textbf{0.509} & \textbf{2.781} & \underline{330.2} & 16.16 \\
\midrule
\multicolumn{11}{@{}l}{\textit{SOTOPIA-All ($n=450$)} --- Matched-policy self-play}\\
Initial & 4.576 & -0.107 & 3.738 & 7.869 & -0.396 & -0.624 & 0.144 & 2.171 & 304.4 & 13.38 \\
Concise prompt & 4.771 & 0.051 & 3.580 & 7.896 & -0.269 & -0.438 & 0.296 & 2.270 & \textbf{208.2} & \underline{11.80} \\
Vanilla OPD & \underline{5.251} & \underline{0.944} & \textbf{4.140} & \textbf{8.364} & \underline{-0.196} & \textbf{-0.200} & \underline{0.349} & \underline{2.665} & \underline{253.0} & \textbf{9.61} \\
TACT & \textbf{6.031} & \textbf{1.260} & \underline{4.036} & \underline{8.320} & \textbf{-0.122} & \underline{-0.269} & \textbf{0.567} & \textbf{2.832} & 313.5 & 14.83 \\
\midrule
\multicolumn{11}{@{}l}{\textit{SOTOPIA-Hard ($n=70$)} --- Fixed initial Qwen3.5-4B}\\
Initial & 3.457 & -0.829 & 3.414 & 7.729 & -0.471 & -1.057 & -0.200 & 1.720 & 254.9 & 11.26 \\
Concise prompt & \underline{3.857} & -0.986 & 3.386 & 7.471 & \underline{-0.257} & -1.243 & 0.229 & 1.780 & \textbf{186.0} & \underline{10.23} \\
Vanilla OPD & 3.686 & \underline{-0.586} & \underline{3.614} & \textbf{8.329} & -0.300 & \underline{-0.629} & \underline{0.243} & \underline{2.051} & \underline{201.1} & \textbf{8.03} \\
TACT & \textbf{4.371} & \textbf{-0.486} & \textbf{3.657} & \underline{8.143} & \textbf{-0.200} & \textbf{-0.571} & \textbf{0.471} & \textbf{2.198} & 233.7 & 11.67 \\
\midrule
\multicolumn{11}{@{}l}{\textit{SOTOPIA-Hard ($n=70$)} --- Fixed Llama-3.1-8B-Instruct}\\
Initial & 3.400 & -1.329 & 3.171 & 7.457 & -0.543 & -1.271 & -0.243 & 1.520 & 282.6 & \underline{11.34} \\
Concise prompt & \underline{3.586} & -1.129 & 3.300 & 7.414 & \textbf{-0.114} & -1.114 & -0.443 & 1.643 & \textbf{203.1} & 11.36 \\
Vanilla OPD & 3.371 & \underline{-0.857} & \textbf{3.671} & \textbf{8.157} & -0.243 & \textbf{-0.671} & \textbf{0.371} & \underline{1.971} & \underline{259.5} & \textbf{9.53} \\
TACT & \textbf{3.900} & \textbf{-0.514} & \underline{3.457} & \underline{8.143} & \underline{-0.157} & \underline{-0.686} & \underline{-0.014} & \textbf{2.018} & 268.3 & 13.40 \\
\midrule
\multicolumn{11}{@{}l}{\textit{SOTOPIA-Hard ($n=70$)} --- Matched-policy self-play}\\
Initial & 3.729 & -0.957 & 3.243 & 7.557 & -0.414 & -1.157 & -0.171 & 1.690 & 247.9 & 10.84 \\
Concise prompt & \underline{3.843} & -0.900 & 3.314 & 7.457 & -0.271 & -0.929 & \underline{0.171} & 1.812 & \textbf{176.4} & \underline{9.84} \\
Vanilla OPD & 2.986 & \underline{-0.657} & \textbf{3.857} & \textbf{8.114} & \textbf{-0.114} & \textbf{-0.429} & 0.057 & \underline{1.973} & \underline{192.5} & \textbf{6.97} \\
TACT & \textbf{4.214} & \textbf{-0.443} & \underline{3.371} & \underline{7.957} & \underline{-0.229} & \underline{-0.886} & \textbf{0.457} & \textbf{2.063} & 237.7 & 11.84 \\
\bottomrule
\end{tabular*}
\end{table}

\paragraph{Partner comparison protocol.}
This comparison tests how the same target policies perform as their interaction
partner changes. The fixed-initial condition reproduces the original main-table
dialogues with the initial Qwen3.5-4B partner. The Llama condition uses a common
Llama-3.1-8B-Instruct partner for all four methods. In self-play, both participants
use the evaluated method, including concise prompting on both roles. Thus,
self-play changes the partner policy across methods, while each fixed-partner
condition holds it constant. Each participant retains its own private
role-visible context. All conditions share the scenario, ordered
character-combination, and predetermined target-role manifest. TACT always uses
the selected 2,970-node checkpoint. DeepSeek-v4-pro at temperature 0 scores the
predetermined target agent, and every method--partner panel contains 450 valid
ratings, including 70 Hard cases. Initial with the fixed-initial partner and
Initial self-play use equivalent policy pairings but separately generated
dialogues, so their difference also illustrates generation and evaluation
variation.

\paragraph{Performance across partners.}
TACT has the highest Goal mean among the four methods in all six
partner--subset blocks (Table~\ref{tab:partner-comparison}). In the order
fixed initial, Llama, and self-play, its All Goal is 5.611/5.807/6.031 and
Hard Goal is 4.371/3.900/4.214. Its Goal advantage over vanilla OPD persists,
but the margin and absolute scores depend on the partner. Communication costs
also change: TACT uses 280.6/330.2/313.5 target tokens and 14.22/16.16/14.83
turns on All. Relative to vanilla OPD, it uses fewer tokens with Llama but
more with the initial partner and in self-play, and more turns in all three
conditions. The comparison therefore supports a consistent Goal ranking across
these evaluated partner settings, alongside partner-dependent score and cost
trade-offs. These descriptive single-run panels do not isolate partner effects
from sampling variation or differences in the recorded recovery procedures.

\paragraph{Llama recovery protocol.}
The two participants retain independent role-visible contexts. Llama uses JSON-schema constrained action generation under the same rule for all four target methods. The initial pass yielded 1,765 valid ratings; eight completed dialogues required judge-only recovery. We then recovered 25 target-action failures through exact-content serialization repair, preserving the successful dialogue prefixes. For the final two failures (one Initial and one vanilla OPD), the original target model generated one replacement action with a legal-action JSON constraint, using the same prompt and seed; subsequent turns used the original decoding settings. These two replacements are recorded separately from format-only repair. No existing valid rating was overwritten, and no hard-coded no-op fallback was used. Target token costs include both discarded original outputs and replacement generations.

\paragraph{Self-play recovery protocol.}
The first pass provided 1,300 valid baseline ratings and 426 valid TACT ratings. Recovery filled 22, 18, 10, and 24 missing cases for Initial, concise prompting, vanilla OPD, and TACT, respectively. Existing valid ratings and successful dialogue prefixes were retained. Completed dialogues needed only bounded judge retries or schema-only clarification; generation failures used the original model and seed, with Kimi restricted to exact-content serialization repair. When content could not be recovered, the recorded self-play protocol executed an empty \texttt{none} action and continued. Exact-content repairs numbered 7/7/7/12 and empty-action fallbacks 7/11/3/11, respectively; these are action-event counts, not dialogue counts. TACT's final case resumed after a technical timeout without a returned response. Costs count all returned raw target tokens, including invalid outputs used in recovery; unreturned computation during timed-out requests is unavailable and is not imputed. These recovery-inclusive results should not be read as failure-free generation.

\paragraph{Comparison protocol.}
The experiment tests whether method-level conclusions persist when the automated judge changes. DeepSeek-v4-pro, Kimi K2.6, and GLM-5.2 assess identical saved interactions from Initial, concise prompting, vanilla OPD, and TACT. The fixed partner is the initial Qwen3.5-4B model; all four methods share the scenario, ordered character-combination, and predetermined target-role manifest. TACT uses the selected 2,970-node checkpoint. Initial uses the original 428 dialogues plus 22 recovered dialogues, excluding its separate historical rerun. No dialogue is regenerated for this comparison, and target token and turn costs remain those in Table~\ref{tab:main-results}.

Kimi and GLM receive the same method-blind rubric and dialogue-history messages, without previous judge answers. For cases with DeepSeek schema-only clarification, these messages reconstruct the original rubric request without the clarification suffix. Kimi and GLM use non-thinking mode at temperature 0.6; DeepSeek uses temperature 0. Matching numerical temperatures does not imply equivalent randomness across providers. Both agents and all seven dimensions must parse successfully; only the predetermined target agent is reported. Bounded transport and invalid-output retries retain the first valid rating. Each judge has 1,800 valid ratings, comprising 450 per method and 70 Hard cases within each panel.

\paragraph{Consistency and disagreement.}
All three judges rank TACT highest in Goal on both All and Hard (Table~\ref{tab:judge-validation}). They also agree on the full All Goal ordering: TACT, vanilla OPD, concise prompting, then Initial. Absolute values shift: TACT receives All Goal 5.611/4.191/5.251 and Hard Goal 4.371/3.171/4.314 from DeepSeek/Kimi/GLM. On Hard, DeepSeek and Kimi rank concise prompting above vanilla OPD and Initial, whereas GLM ranks Initial above concise prompting and vanilla OPD. TACT leads Avg on All under every judge, and on Hard under DeepSeek and GLM; Kimi slightly favors vanilla OPD on Hard Avg (1.255 versus 1.249). Thus, the primary Goal conclusion persists across these judges, while absolute scores, lower-ranked methods, and individual dimensions are judge-sensitive. This descriptive comparison does not establish statistical significance, stability under repeated judge sampling, or agreement with human assessments.

\clearpage
\begin{table}[t]
\centering\footnotesize\setlength{\tabcolsep}{2.5pt}
\caption{\textbf{Full cross-judge score panels.} Seven native-scale SOTOPIA dimensions and their arithmetic mean for the identical dialogues in Table~\ref{tab:judge-validation}. Best and second-best displayed means are marked separately within each judge--subset block; markings do not imply statistical significance.}
\label{tab:judge-dimensions}
\begin{tabular*}{\linewidth}{@{\extracolsep{\fill}}lrrrrrrrr@{}}
\toprule
Method & Goal $\uparrow$ & Rel. $\uparrow$ & Kno. $\uparrow$ & Bel. $\uparrow$ & Sec. $\uparrow$ & Rules $\uparrow$ & Fin. $\uparrow$ & Avg $\uparrow$ \\
\midrule
\multicolumn{9}{@{}l}{\textit{SOTOPIA-All ($n=450$)} --- DeepSeek-v4-pro}\\
Initial & 4.327 & -0.193 & 3.707 & 7.800 & -0.442 & -0.669 & 0.102 & 2.090 \\
Concise prompt & 4.691 & -0.009 & 3.687 & 7.827 & -0.311 & -0.476 & 0.216 & 2.232 \\
Vanilla OPD & \underline{4.960} & \underline{0.580} & \underline{4.060} & \textbf{8.376} & \textbf{-0.171} & \textbf{-0.224} & \underline{0.307} & \underline{2.555} \\
TACT & \textbf{5.611} & \textbf{0.818} & \textbf{4.171} & \underline{8.356} & \underline{-0.278} & \underline{-0.229} & \textbf{0.527} & \textbf{2.711} \\
\midrule
\multicolumn{9}{@{}l}{\textit{SOTOPIA-All ($n=450$)} --- Kimi K2.6}\\
Initial & 3.080 & -0.953 & \underline{3.144} & 5.924 & -0.749 & -1.280 & -0.687 & 1.211 \\
Concise prompt & 3.340 & -0.791 & 3.118 & 5.784 & -0.489 & -1.149 & -0.418 & 1.342 \\
Vanilla OPD & \underline{3.636} & \underline{0.011} & \textbf{3.538} & \textbf{6.329} & \textbf{-0.256} & \textbf{-0.520} & \underline{-0.360} & \underline{1.768} \\
TACT & \textbf{4.191} & \textbf{0.207} & \textbf{3.538} & \underline{6.171} & \underline{-0.369} & \underline{-0.671} & \textbf{0.049} & \textbf{1.874} \\
\midrule
\multicolumn{9}{@{}l}{\textit{SOTOPIA-All ($n=450$)} --- GLM-5.2}\\
Initial & 4.244 & -0.202 & 3.904 & 7.038 & -0.773 & -0.764 & 0.160 & 1.944 \\
Concise prompt & 4.404 & -0.071 & 3.922 & 7.024 & -0.540 & -0.693 & 0.320 & 2.052 \\
Vanilla OPD & \underline{4.769} & \underline{0.553} & \underline{4.258} & \textbf{7.347} & \textbf{-0.311} & \textbf{-0.364} & \underline{0.336} & \underline{2.370} \\
TACT & \textbf{5.251} & \textbf{0.800} & \textbf{4.282} & \underline{7.311} & \underline{-0.396} & \underline{-0.367} & \textbf{0.618} & \textbf{2.500} \\
\midrule
\multicolumn{9}{@{}l}{\textit{SOTOPIA-Hard ($n=70$)} --- DeepSeek-v4-pro}\\
Initial & 3.457 & -0.829 & 3.414 & 7.729 & -0.471 & -1.057 & -0.200 & 1.720 \\
Concise prompt & \underline{3.857} & -0.986 & 3.386 & 7.471 & \underline{-0.257} & -1.243 & 0.229 & 1.780 \\
Vanilla OPD & 3.686 & \underline{-0.586} & \underline{3.614} & \textbf{8.329} & -0.300 & \underline{-0.629} & \underline{0.243} & \underline{2.051} \\
TACT & \textbf{4.371} & \textbf{-0.486} & \textbf{3.657} & \underline{8.143} & \textbf{-0.200} & \textbf{-0.571} & \textbf{0.471} & \textbf{2.198} \\
\midrule
\multicolumn{9}{@{}l}{\textit{SOTOPIA-Hard ($n=70$)} --- Kimi K2.6}\\
Initial & 2.400 & -1.543 & 2.871 & 5.843 & -0.571 & -1.757 & -1.271 & 0.853 \\
Concise prompt & \underline{2.686} & -1.957 & 2.643 & 5.614 & -0.343 & -2.000 & -1.057 & 0.798 \\
Vanilla OPD & 2.529 & \textbf{-1.129} & \textbf{3.186} & \textbf{6.314} & \textbf{-0.271} & \textbf{-1.071} & \underline{-0.771} & \textbf{1.255} \\
TACT & \textbf{3.171} & \underline{-1.243} & \underline{3.071} & \underline{6.029} & \underline{-0.329} & \underline{-1.457} & \textbf{-0.500} & \underline{1.249} \\
\midrule
\multicolumn{9}{@{}l}{\textit{SOTOPIA-Hard ($n=70$)} --- GLM-5.2}\\
Initial & \underline{3.643} & -0.657 & 3.957 & 7.243 & \underline{-0.229} & -0.871 & 0.000 & 1.869 \\
Concise prompt & 3.557 & -1.014 & 3.800 & 7.214 & -0.443 & -1.257 & 0.229 & 1.727 \\
Vanilla OPD & 3.429 & \underline{-0.429} & \underline{4.029} & \textbf{7.571} & -0.271 & \underline{-0.743} & \underline{0.357} & \underline{1.992} \\
TACT & \textbf{4.314} & \textbf{-0.286} & \textbf{4.200} & \underline{7.543} & \textbf{-0.143} & \textbf{-0.714} & \textbf{0.800} & \textbf{2.245} \\
\bottomrule
\end{tabular*}
\end{table}

\subsection{Qualitative Analysis}
Case analysis should include productive expression edits, strategy changes that resolve a disagreement, and failures where local compression omits conditions or transfers effort to later turns. Cases with improved Goal but worse relationship or constraint adherence are particularly relevant. Each empirical case requires a traceable prefix, original and candidate actions, selection evidence, and observed subsequent interaction. Illustrative dialogues must be marked as illustrations rather than experimental observations; no additional empirical cases are asserted here.

\end{document}